\pdfoutput=1
\documentclass[runningheads]{llncs}
\usepackage{graphicx}
\usepackage{comment}
\usepackage{epsfig}
\usepackage{booktabs}
\usepackage{amsmath,amssymb} 
\usepackage{color}

\usepackage{amsfonts}       
\usepackage{nicefrac}       
\usepackage{color}
\usepackage{multirow}
\usepackage[super]{nth}
\usepackage{subfig}
\usepackage[normalem]{ulem} 
\usepackage{multirow, makecell}
\usepackage{xspace}
\usepackage[pagebackref=true,breaklinks=true,letterpaper=true,colorlinks,bookmarks=false]{hyperref}

\newcommand{\myparagraph}[1]{\vspace{5pt}\noindent{\bf #1}}

\newcommand{\papername}{\textbf{\textsc{VisualCOMET}}}

\newcommand{\vcg}{Visual Commonsense Graphs }

\newcommand{\numinf}{1.4 million}

\usepackage[width=122mm,left=12mm,paperwidth=146mm,height=193mm,top=12mm,paperheight=217mm]{geometry}

\begin{document}
\pagestyle{headings}
\mainmatter
\def\ECCVSubNumber{3684}  
\title{VisualCOMET: Reasoning about \\ the Dynamic Context of a Still Image \\ \href{https://visualcomet.xyz}{visualcomet.xyz}\vspace{-1em}}


%
\titlerunning{VisualCOMET}
\author{Jae Sung Park$^{1,2}$, Chandra Bhagavatula$^{2}$, Roozbeh Mottaghi$^{1,2}$, \\ Ali Farhadi$^{1}$, Yejin Choi $^{1,2}$}

\authorrunning{Park et al.}

\institute{Paul G. Allen School of Computer Science \& Engineering, WA, USA \and
Allen Institute for Artificial Intelligence, WA, USA}

\maketitle

\setcounter{footnote}{0}

\vspace{-0.7cm}
\begin{figure}[h!]
\scriptsize
\begin{center}
\includegraphics[width=0.8\linewidth]{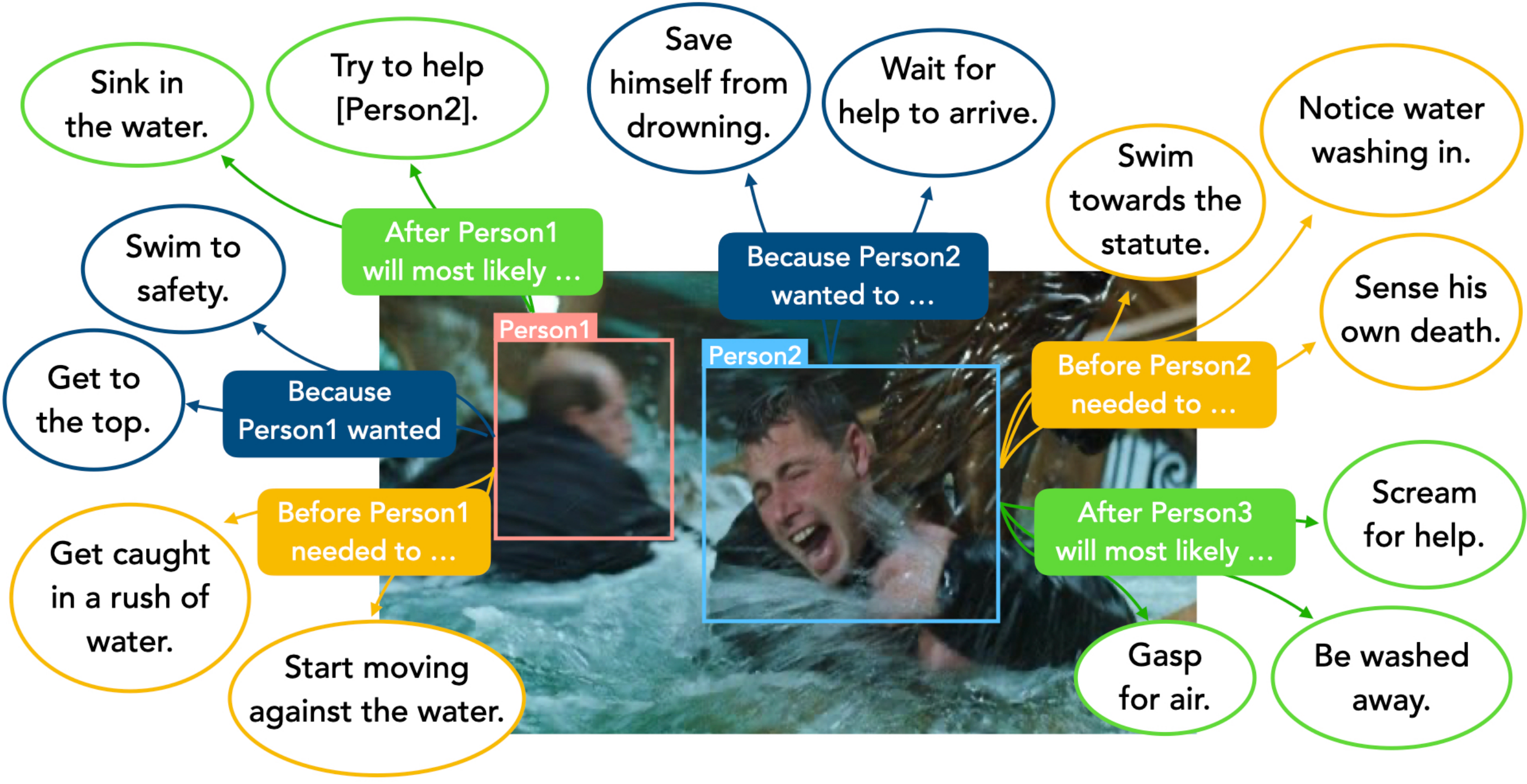} 
\caption{\footnotesize Given a person in the image, \papername{}  provides a \emph{graph} of common sense inferences about 1) what needed to happen before, 2) intents of the people at present, and 3) what will happen next.}
\label{fig:teaser}
\end{center}
\vspace{-1.3cm}
\end{figure}

\begin{abstract}
Even from a single frame of a still image, people can reason about the dynamic story of the image \emph{before}, \emph{after}, and \emph{beyond} the frame. For example, given an image of a man struggling to stay afloat in water, we can reason that the man fell into the water sometime in the past, the intent of that man at the moment is to stay alive, and he will need help in the near future or else he will get washed away.
We propose \papername,\footnote{\textbf{Visual} \textbf{Com}monsense R\textbf{e}asoning in \textbf{T}ime.} the novel framework of visual commonsense reasoning tasks to predict events that might have happened before, events that might happen next, and the intents of the people at present. 
To support research toward visual commonsense reasoning, we introduce the first large-scale repository of \textbf{\vcg} that consists of over \textbf{\numinf} textual descriptions of visual commonsense inferences carefully annotated over a diverse set of 59,000 images, each paired with short video summaries of before and after. 
In addition, we provide person-grounding (i.e., co-reference links) between  
people appearing in the image and people mentioned in the textual commonsense descriptions, allowing for tighter integration between images and text. 
We establish strong baseline performances on this task and demonstrate that integration between visual and textual commonsense reasoning is the key and wins over non-integrative alternatives.
\end{abstract}

\section{\label{sec:intro} Introduction}


Given a still image, people can reason about the rich dynamic story underlying the visual scene that goes far beyond the frame of the image. For example, in Figure~\ref{fig:teaser}, given the image of a desperate man holding onto a statue in water, we can reason far beyond what are immediately visible in that still frame; sometime in the past, an accident might have happened and a ship he was on might have started sinking. Sometime in the future, he might continue struggling and eventually be washed away. In the current moment, his intent and motivation must be that he wants to save himself from drowning. This type of visual understanding requires a major leap from recognition-level understanding to cognitive-level understanding, going far beyond the scope of image classification, object detection, activity recognition, or image captioning. An image caption such as ``a man in a black shirt swimming in water'', for example, while technically correct, falls far short of understanding the dynamic situation captured in the image that requires reasoning about the context that spans before, after, and beyond the frame of this image. Key to this rich cognitive understanding of visual scenes is visual commonsense reasoning, which in turn, requires rich background knowledge about how the visual world works, and how the social world works.


In this paper, we propose \papername, a new framework of task formulations to reason about the rich visual context that goes beyond the immediately visible content of the image, ranging from events that might have happened before, to events that might happen next, and to the intents of the people at present. 
To support research toward visual commonsense reasoning, we introduce the first  large-scale repository of \textbf{\vcg} that consists of \textbf{\numinf} textual descriptions of visual commonsense inferences that are carefully annotated over a diverse set of about 59,000 people-centric images from VCR \cite{zellers19cvpr}. In addition, we provide person-grounding (i.e., co-reference links) between people appearing in the image and people mentioned in the textual commonsense descriptions, allowing for tighter integration between images and text. 
The resulting Visual Commonsense Graphs are rich, enabling a number of task formulations with varying levels of difficulties for future research. 


We establish strong baseline performances on such tasks based on GPT-2  transformer architecture \cite{radford2019language} to combine visual and textual information. Quantitative results and human evaluation show that integrating both the visual and textual commonsense reasoning is the key for enhanced performance. Furthermore, when the present eventual description is not available and only image is given, we find that the model trained to predict both events and inferential sentences performs better than the one trained to predict only inferences.

In summary, our contributions are as follows. (1) We introduce a new task of visual commonsense reasoning for cognitive visual scene understanding, to reason about events before and after and people's intents at present. (2) We present the first large-scale repository of Visual Commonsense Graphs that contains more than 1M textual descriptions of commonsense inferences over 60K complex visual scenes. (3) We extend the GPT-2 model to incorporate visual information and allow direct supervision for grounding people in images.  (4) Empirical results and human evaluations show that model trained jointly with visual and textual cues outperform models with single modality, and can generate meaningful inferences from still images.
\section{\label{sec:related} Related Work}

\myparagraph{Visual Understanding with Language:} Various tasks have been introduced for joint understanding of visual information and language, such as image captioning \cite{Chen2015MicrosoftCC,vinyals2015show,Sharma2018ConceptualCA}, visual question answering \cite{Agrawal2015VQAVQ,Johnson2016CLEVRAD,okvqa} and referring expressions \cite{kazemzadeh-etal-2014-referitgame,Plummer2015Flickr30kEC,mao2106}. These works, however, perform inference about only the current content of images and fall short of understanding the dynamic situation captured in the image, which is the main motivation of our work. There is also a recent body of work addressing representation learning using vision and language cues \cite{tan2019lxmert,Lu2019ViLBERTPT,su2019vl}. We propose a baseline for our task, which is inspired by these techniques. 

\myparagraph{Visual Commonsense Inference:} 
Prior works have tried to incorporate commonsense knowledge in the context of visual understanding. \cite{Vedantam2015LearningCS} use human-generated abstract scenes made from clipart to learn common sense, but not on real images. \cite{Pirsiavash2014InferringTW} try to infer the motivation behind the actions of people from images. Visual Commonsense Reasoning (VCR) \cite{zellers19cvpr} tests if the model can answer questions with rationale using commonsense knowledge. While \cite{zellers19cvpr} includes rich visual common sense information, their question answering setup makes it difficult to have models to generate commonsense inferences. ATOMIC \cite{sap19aaai} provides a commonsense knowledge graph containing if-then inferential textual descriptions in generative setting; however, it relies on generic, textual events and does not consider visually contextualized information. In this work, we are interested in extending \cite{zellers19cvpr} and \cite{sap19aaai} for general visual commonsense by building a large-scale repository of visual commonsense  graphs and models that can explicitly generate commonsense inferences for given images.

\myparagraph{Visual Future Prediction:}
There is a large body of work on future prediction in different contexts such as future frame generation \cite{Ranzato2014VideoM,pmlr-v37-srivastava15,xue16,vondrick16,Mathieu15,villegas19,castrejon19}, prediction of the trajectories of people and objects \cite{walker14,Alahi_2016_CVPR,mottaghi16b}, predicting human pose in future frames \cite{fragkiadaki15,Walker_2017_ICCV,chao17} and semantic future action recognition \cite{lan14,zhou15,chen19}. In contrast to all
these approaches, we provide a compact description for the future events using language.

\section{\label{sec:task}Task: Cognitive Image Understanding via \\ Visual Commonsense Graphs}

\begin{figure}[t]
\scriptsize
\begin{center}
\includegraphics[width=\linewidth]{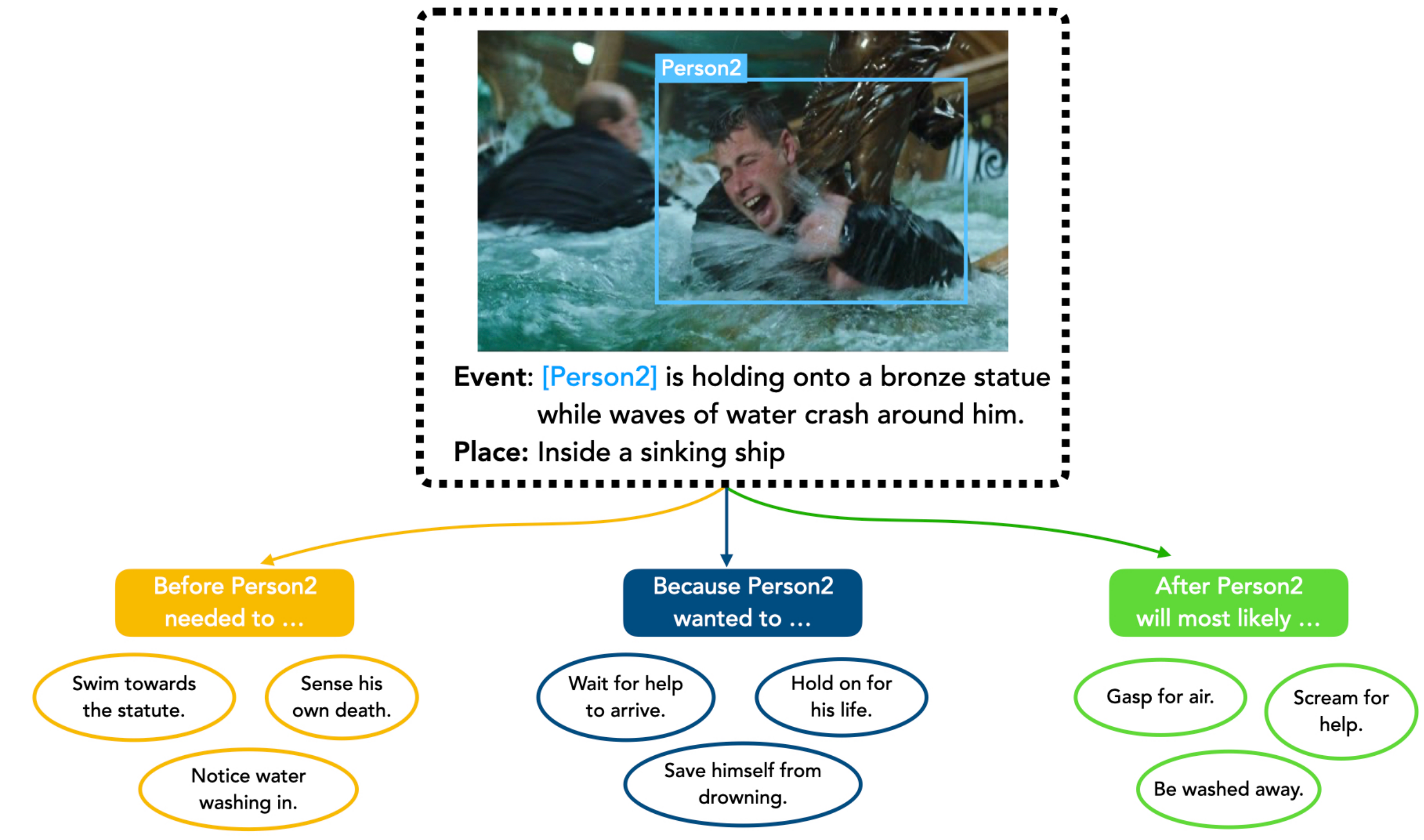} \\
\caption{\textbf{Task Overview:} Our proposed task is to generate commonsense inferences of \textbf{events before}, \textbf{events after} and \textbf{intents at present}, given an image, a description of an \textbf{event at present} in the image and a plausible scene / location of the image.}
\label{fig:task}
\end{center}
\end{figure}

\subsection{Definition of Visual Commonsense Graphs}
The ultimate goal is to generate the entire visual commonsense graph illustrated in Figure~\ref{fig:teaser} that requires reasoning about the dynamic story underlying the input image. This graph consists of four major components: 
\begin{itemize}
\item
(1) a set of textual descriptions of \textbf{events at present}, 
\item (2) a set of commonsense inferences on \textbf{events before}, 
\item (3) a set of commonsense inferences on \textbf{events after}, and 
\item (4) a set of commonsense inferences on people's \textbf{intents at present}.  
\end{itemize}

The events before and after can broadly include any of the following: (a) \emph{actions} people might take before and after (e.g., people jumping to the water), (b) \emph{events} that might happen before and after (e.g., a ship sinking), and (c) \emph{mental states} of people before and after (e.g., people scared and tired). Our design of the commonsense graph representation is inspired by ATOMIC \cite{sap19aaai}, a text-only atlas of machine commonsense knowledge for \emph{if-then} reasoning, but tailored specifically for cognitive understanding of visual scenes in images.

\myparagraph{Location and Person Grounding:}
In addition, the current event descriptions are accompanied by additional textual descriptions of the \textbf{place} or the overall scene of the image, e.g., ``at a bar'' or ``at a party''. We also provide  person-grounding  (i.e.,  co-reference  links)  between  people  appearing in the image and people mentioned in the textual commonsense descriptions,  allowing  for  tighter  integration  between  images  and  text.

\myparagraph{Dense Event Annotations with Visual Commonsense Reasoning:}
Generally speaking, the first component of the visual commonsense graph, ``\emph{events at present}'', is analogous to dense image captioning in that it focuses on the immediate visual content of the image, while components (2) - (4), \emph{events before and after} and \emph{intents at present}, correspond to visual commonsense reasoning.

Importantly, in an image that depicts a complex social scene involving multiple people engaged in different activities simultaneously, the inferences about before, after, and intents can be ambiguous as to which exact current event the inferences are based upon. Therefore, in our graph representation, we link up all the commonsense inferences to a specific event at present. 

\subsection{Definition of Tasks}
Given the complete visual commonsense graph representing an image, we can consider multiple task formulations of varying degrees of difficulties. In this paper, we focus on two such tasks: (1) Given an image and one of the events at present, the task is to generate the rest of visual commonsense graph that is connected to the specific current event. (2) Given an image, the task is to generate the complete set of commonsense inferences from scratch. 

\section{\label{sec:dataset}Dataset Overview}

\begin{figure}[ht!]%
    \centering
    \includegraphics[width=30pc]{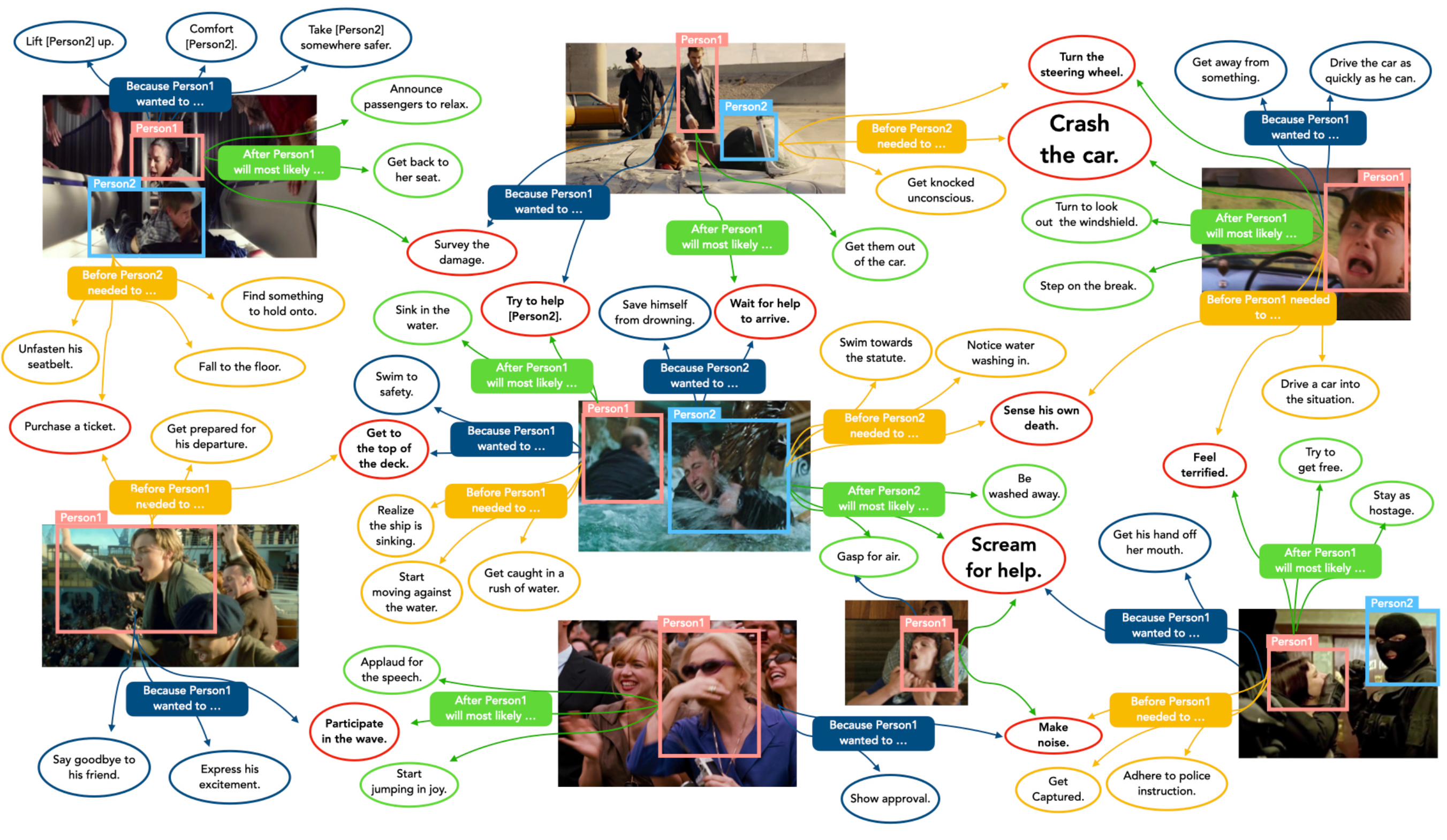} \\ 
    \caption{Overview of our \textbf{Visual Commonsense Graphs}. We see that a diverse set of images are covered and connected with inference sentences. Red bubbles indicate if inference sentences shared by two or more images.}
    \label{fig:v_comet}
\end{figure}

We present the first large-scale dataset of Visual Commonsense Graphs for images with person grounding (i.e., multimodal co-reference chains). We collect a dataset of \numinf{} commonsense inferences over 59,356 images and 139,377 distinct {events at present} (Table \ref{tab:stats}). Figure \ref{fig:v_comet} gives an overview of our Visual Commonsense Graphs including a diverse set of images, connected with the inference sentences \footnote{Larger figure available in the Appendix.}.

\begin {table}[!ht]
\begin{center}
\setlength\tabcolsep{0pt} 
\begin{tabular}{@{}l@{}c@{\ \ }||@{\ }c@{ \ \ }@{}c@{\ \ }@{}c@{\ \ }|@{\ }c@{\ \ }}
\toprule
\textbf{} & \textbf{       } & \textbf{  Train     } & \textbf{    Dev  } &  \textbf{  Test  } & \textbf{ Total }\\
\midrule
\# Images/Places & & 47,595 & 5,973 & 5,968 & \bf 59,356\\
\# Events at Present & & 111,796 & 13,768 & 13,813 & \bf 139,377 \\
\midrule 
\# Inferences on Events Before & & 467,025 & 58,773 & 58,413& 584,211\\
\# Inferences on Events After & & 469,430 & 58,665 & 58,323 & 586,418\\
\# Inferences on Intents at Present & & 237,608 & 28,904 & 28,568 & 295,080\\
\midrule
\# Total Inferences & & 1,174,063 & 146,332 & 145,309 &\bf 1,465,704\\
\bottomrule
\end{tabular}
\end{center}
\caption{\textbf{Statistics} of our Visual Commonsense Graph repository: there are in total 139,377 distinct Visual Commonsense Graphs over 59,356 images involving 1,465,704 commonsense inferences.} 
\label{tab:stats}
\end {table}


\subsection{Source of Images}
As the source of the images, we use the VCR \cite{zellers19cvpr} dataset that consists of images corresponding to complex visual scenes with multiple people and activities. The dataset also includes automatically detected object bounding boxes, and each person in the image uniquely identified with a referent tag (e.g. Person1 and Person2 in Fig \ref{fig:teaser}). 


\subsection{Crowdsourcing Visual Commonsense Graphs}
\label{sec:crowd}

\begin{figure}[t]
\scriptsize
\begin{center}
\includegraphics[width=\linewidth]{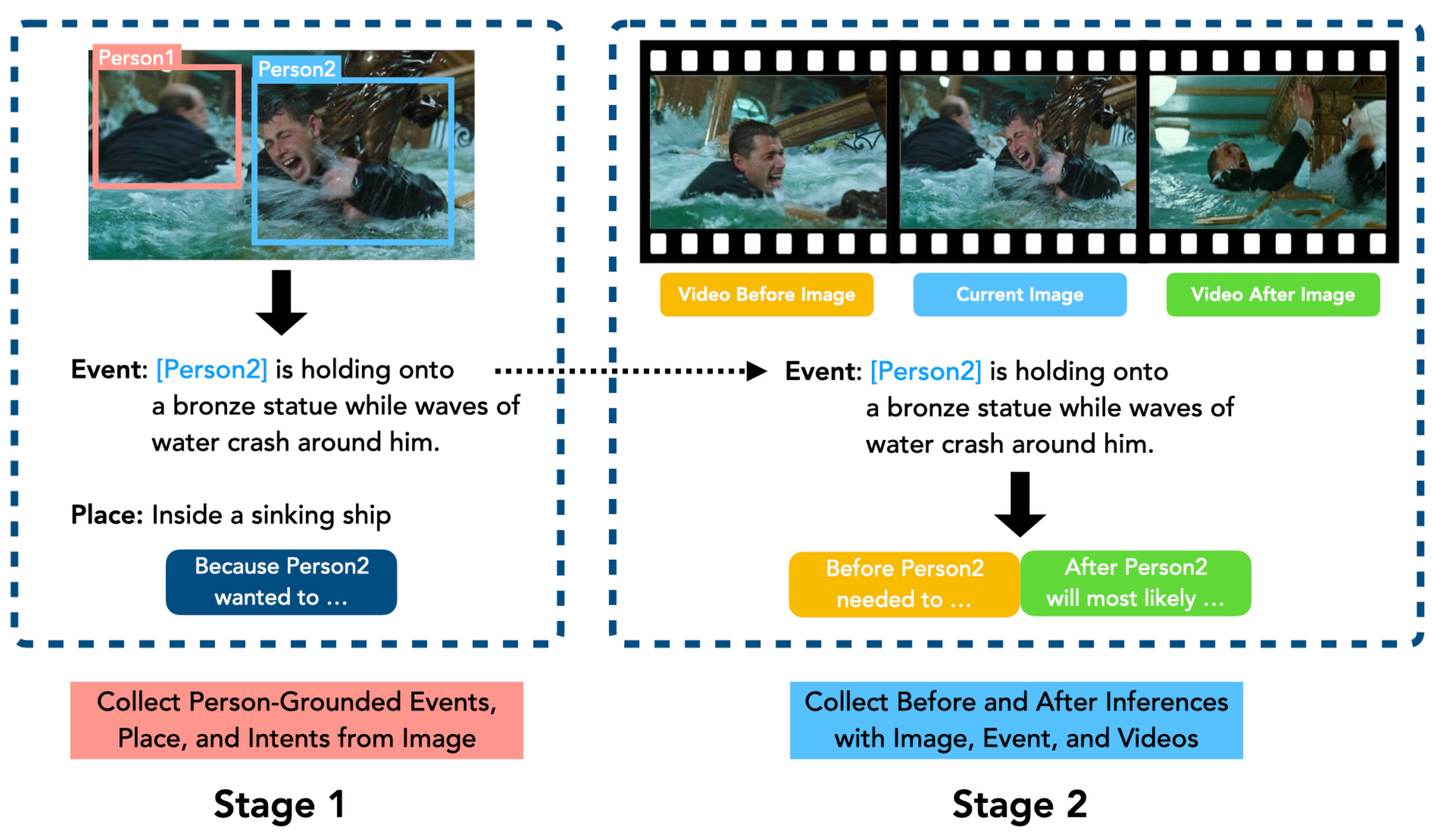} \\
\caption{\textbf{Annotation Pipeline:} Our two-stage crowdsourcing annotation pipeline used for collecting our high-quality Visual Commonsense Graphs.}
\label{fig:annotation}
\end{center}
\end{figure}

Annotating the entire commonsense graph solely from an image is a daunting task even for humans.
We design a two-stage crowdsourcing pipeline to make the annotation task feasible and to obtain focused and consistent annotations. We run our annotation pipeline on Amazon Mechanical Turk (AMT) platform and maintain the ethical pay rate of at least \$15/hr. This amounts to $\$4$ per image on average.
Figure \ref{fig:annotation} shows an overview of our annotation pipeline.

\myparagraph{Stage 1: Grounded Event Descriptions with Locations and Intents}



In the first stage, we show crowdworkers an image along with tags  identifying each person in the image. Crowdworkers select a person and author a description for the event involving that person. 
One key concern during event annotation is to encourage crowdworkers to annotate informative, interesting events as opposed to low-level events like standing, sitting, looking, etc. While technically correct, such descriptions do not contribute to higher-level understanding of the image. To obtain more meaningful events, we ask crowdworkers to write an event description and intent inferences at the same time. Finally, we ask crowdworkers to annotate the plausible location of the scene depicted in the image. In addition to priming workers, such location information provides more contextualized information for the task. The location information is not just a physical place, but can also include occasions, e.g., in a business meeting. At the end of stage 1, we collect (i) the location of an image, (ii) two to three events for each image, and (iii) two to four intents for each event of each image.




\myparagraph{Stage 2: Collecting Before and After Inferences}

In stage 2, we collect visual commonsense inferences of what might have happened \textit{before} and what might happen \textit{after} for each event description for each image annotated in stage 1 above. Images in our dataset were originally part of movie scenes. Based on the timestamp of the image being annotated, we show crowdworkers two short, fast-forwarded clips of events that happen before and after the image. This allows crowdworkers to author inferences that are more meaningful, rather than authoring correct but trivial inferences -- e.g. ``before, Person1 needed to be born", ``after, Person1 will be dead", etc. 

We assign two workers for each event and ask each to annotate between two and four \textit{before} and \textit{after} inferences. 
At the end of the two stages in our annotation pipeline, we have up to ten (2 intent, 4 before, 4 after) inferences for each pair of image and a textual description of event at present.

\section{\label{sec:approach} Our Approach}

Our task assumes the following inputs for each image: a sequence of visual embeddings $\mathcal{V}$ representing the image and people detected in the image, grounded event description $e$, scene's location information $p$, and inference type $r$. Then, we wish to generate a set of possible inferences $H = \{s_1^r, s_2^r, ... s_{|H|}^r\}$.

\subsection{Visual Features}
The sequence of visual representations $\mathcal{V}$ consists of a representation of the whole image and an additional representations for each person detected in the image. We use Region of Interest (RoI) Align features \cite{He2017} from Faster RCNN \cite{ren2015faster} as our visual embedding and pass it through a non-linear layer to obtain the final representation for an image or each detected person. The final sequence of representations $\mathcal{V} = \{v_0, v_1, .. v_k \}$ where $k$ is the number of people detected.

As described in \S\ref{sec:crowd}, we provide special tags identifying each person in the image (e.g. Person1 in Fig. \ref{fig:annotation}) in our dataset.  
To use these tags, we introduce new person tokens, e.g. [Person1], in the vocabulary and create additional word embedding for these tokens. 
Then, we sum the visual representation for a person with the word embedding of the token referencing the person in text. This way, our model has visually grounded information about the image. We refer to this approach as ``Person Grounding" (PG) input. 


\subsection{Text Representation}
 Transformer models used for language tasks \cite{devlin2018bert,radford2019language} use special separator tokens to enable better understanding of the input structure. Since our task involves textual information of different kinds (event, place, and relation), we follow \cite{bosselut19acl,Zellers2019DefendingAN,Bhagavatula2019AbductiveCR} to include special tokens for our language representation as well. Specifically, we append special token indicating the start and end of image (e.g. \texttt{s\_img}, \texttt{e\_img}), event, place, and inference fields. To generate inference statements, we use one of the three inference types (\textit{before}, \textit{intent}, \textit{after}) as the start token, depending on the desired dimension.
 
\begin{figure}[!ht]
\scriptsize
\begin{center}
\includegraphics[width=\linewidth]{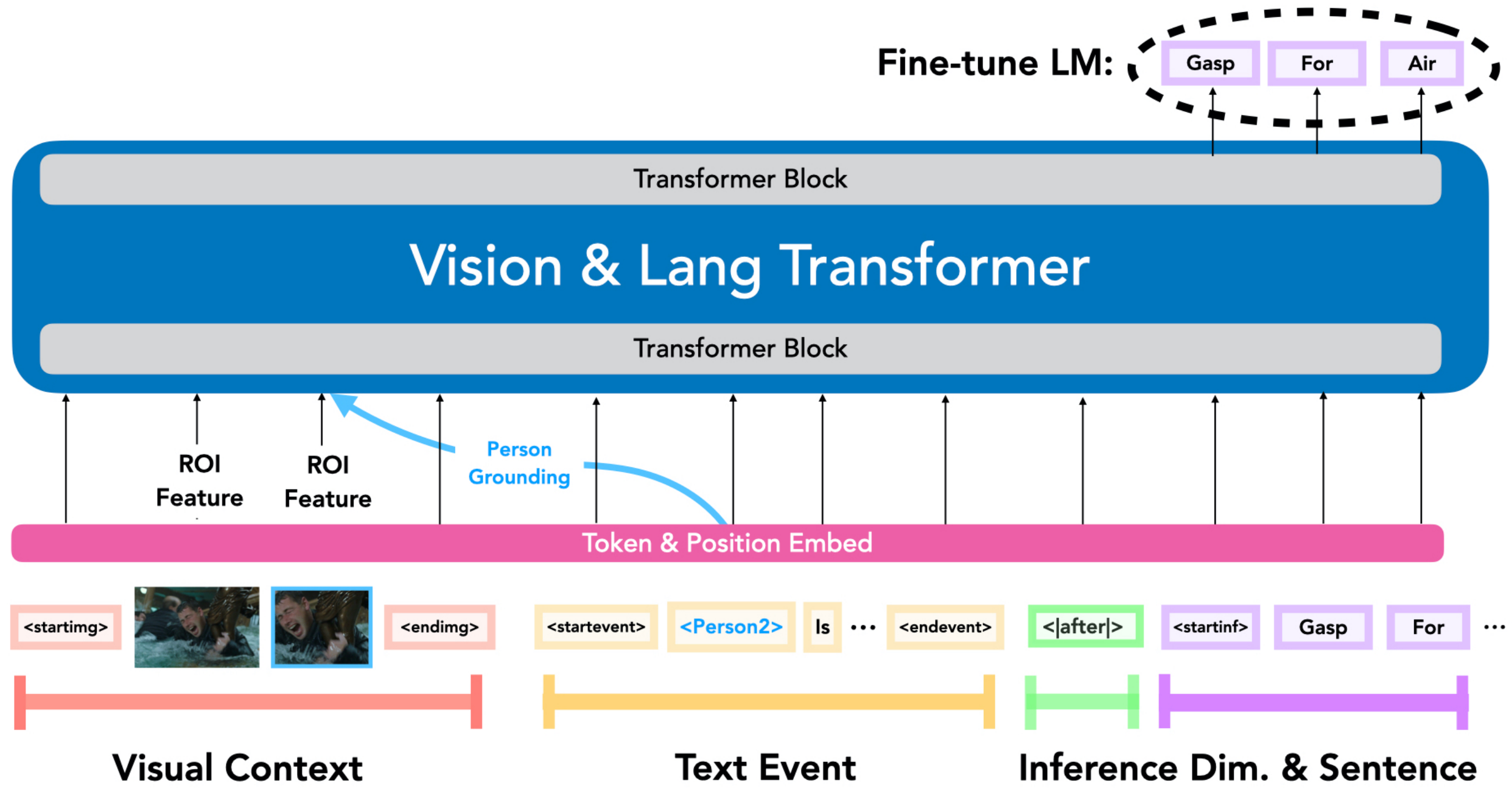} \\
\caption{\textbf{Model Overview.} Vision-Language Transformer for our approach. Our sequence of inputs uses special tokens indicating the start and end of image, event, place, and inference. We only show the start token in the figure for simplicity.} 
\label{fig:model}
\end{center}
\vspace{-0.5cm}
\end{figure}

\subsection{Single Stream Vision-Language Transformer}

We fix the model architecture as GPT-2 \cite{radford2019language}, a strong Transformer model \cite{vaswani17nips} for natural language generation, conditioned on $\mathcal{V},e,p$. 
Our model is a single stream transformer that encodes visual and language representations with a single transformer model, which has been shown to be more effective in vision and language tasks \cite{Chen2019UNITERLU,zhou20aaai} compared to designing separate transformer models for each modality \cite{Lu2019ViLBERTPT}.

For each inference $s_h^r \in H$, our objective is to maximize $P(s_h^r | v,e,p,r)$. Suppose $s_h^r = \{w_{h1}^r, w_{h2}^r, ... w_{hl}^r \}$ is a sequence of $l$ tokens. Then, we minimize the negative log-likelihood loss over inference instances in dataset:

\begin{equation}
\begin{aligned} 
\mathcal{L} = - \sum_{i=1}^l{\log P(w_{hi}^r | w_{h<i}^r, r, e, p, v)}
\end{aligned}
\label{eq:nll_loss}
\end{equation}

While our dataset provides events associated with each image, it is impractical to assume the availability of this information on new images. We experiment with a more general version of our model which does not take $e$ and $p$ as input. Nonetheless, we can supervise such models to generate $e$ and $p$ in the training phase. If we denote the \emph{event at present} $\{e\}= \{w_1^e, w_2^e, ... w_n^e\}$ and place $\{p\}= \{w_1^p, w_2^p, ... w_m^p\}$ as a sequence of tokens, we apply the seq2seq loss on $e, p$ (EP Loss in Section \ref{sec:experiments}) as follows:

\begin{equation}
\begin{aligned} 
\mathcal{L} = & - \sum_{i=1}^n{\log P(w_{i}^e | w_{<i}^e,  v))} - \sum_{i=1}^m{\log P(w_{i}^p | w_{<i}^p, e, v))} \\ & - \sum_{i=1}^l{\log P(w_{hi}^r | w_{h<i}^r,r,e,p,v)}
\end{aligned}
\label{eq:nll_aux_loss}
\end{equation}

\section{\label{sec:experiments} Experiments and Results}

\subsection{Implementation Details}
We use Adam optimizer \cite{kingma2014adam} with a learning rate of 5e-5 and batch size of 64. Visual features for image and person embeddings use ResNet101 \cite{he2016deep} backbone pretrained on ImageNet \cite{deng2009imagenet}. We set the maximum number of visual features to 15. We use pre-trained GPT2-base model \cite{radford2019language} as our model architecture with maximum total sequence length as 256. For decoding, we use nucleus sampling \cite{Holtzman2019TheCC} with $p=0.9$, which has shown to be effective generating text that is diverse and coherent. We have found beam search, which is a popular decoding scheme for generating multiple candidates, to be repetitive and produce uninteresting inferences. We report the effect of different decoding schemes in the supplementary material.

\subsection{Experimental Setup}

\myparagraph{Baselines based on Different Inputs}

In our experiments, we fix the same model architecture but ablate on the inputs available, e.g. place, event, and image. We also measure the effect of Person Grounding (PG) trick stated in Section 5.1. The models are trained with the same seq2seq objective in Eq. \ref{eq:nll_loss}, and we mask out the visual and/or textual input based on the ablation of interest. We additionally experiment if learning to generate the event at present and place can improve the performance of generating the inferences using the objective in Eq. \ref{eq:nll_aux_loss}. For simplicity, we denote the loss on the two textual input as [+ EP. Loss]. Thus, we test two settings when generating the inferences: 1) one that uses event, place, and image, and 2) one that uses only image. We mark the two options in the Text Given column. \footnote{We have tried running inferences on predicted events at present, but have gotten worse results than using no events. We report the results on predicted events in the supplemental.}

\begin {table}[t!]
\begin{center}
\scalebox{0.8}{
\begin{tabular}{@{}l@{}c@{\ \ }c@{\ \ }c@{\ \ }c@{\ \ }c@{\ \ }c@{\ \ }@{}l@{}}
\toprule
\textbf{Modalities} & \textbf{Text }& \textbf{B-2 } & \textbf{M  } & \textbf{C  } & \textbf{Acc@50 } & \textbf{Unique} & \textbf{Novel}\\

\textbf{}  & \textbf{Given}\\

\midrule
Place & Yes & 5.87 & 6.25 & 4.69 & 14.55 & 6.84 & 47.57 \\ 
Event & Yes & 10.99 & 9.58 & 14.81 & 31.95 & 39.56 & 47.19 \\ 
\textbf{Event + Place} & Yes & 11.46 & 9.82 & 15.73 & 33.06 & 41.39 & 47.61 \\ 
\midrule
Image + Place & Yes & 7.42 & 7.33 & 6.69 & 20.39 & 27.70 & 46.50 \\ 
Image + Event & Yes & 12.52 & 10.73 & 16.49 & 37.00 & 42.83 & 47.40\\ 
Image + Event + Place & Yes & 12.78 & 10.87 & 17.12 & 38.25 & 43.83 & 48.15 \\ 
Image + Event + Place + EP Loss & Yes  & 11.15 & 10.02 & 13.60 & 33.23 & 42.13 & 51.25 \\ 
\textbf{Image + Event + Place + PG}& Yes & \textbf{13.50} & \textbf{11.55} & \textbf{18.27} & \textbf{38.72} & \textbf{44.49} & 49.03\\ 
Image + Event + Place + PG + EP Loss & Yes & 12.10 & 10.74 & 15.00 & 34.07 & 42.33 & \textbf{51.73}\\ 
\midrule
No Input & No & 3.76 & 5.23 & 2.07 & 6.87 & 0.00 & 33.33 \\
Image & No & 6.79 & 7.13 & 5.63 & 18.22 & 26.38 & 46.80 \\ 
\textbf{Image + PG} & No & 8.20 & 8.44 & 7.61 & 21.5 & 29.09 & 45.53 \\ 
\midrule
Image + Event + Place & No & 6.97 & 7.55 & 6.01 & 16.81 & 24.75 & 45.27 \\
Image + Event + Place + EP Loss & No & 7.06 & 7.77 & 6.37 & 20.02 & 31.60 & 50.77 \\
Image + Event + Place + PG & No & 8.80 & 9.19 & 8.77 & 17.35 & 27.42 & 47.37 \\
\textbf{Image + Event + Place + PG + EP Loss} & No & \textbf{10.21}  & \textbf{10.66} & \textbf{11.86} & \textbf{22.7} & \textbf{33.90} & \textbf{49.84}\\

\midrule 
GT & - & - & - & - & - & 74.34 & 54.98\\  
\bottomrule
\end{tabular}
}
\end{center}
\caption{\textbf{Ablation Results.} Ablations of our baseline model on the Validation set. We use nucleus sampling with $p=0.9$ to generate 5 sentences for all models. Automatic metrics used are BLEU-2 (B-2) \cite{bleu}, METEOR (M) \cite{meteor}, and CIDER (C) \cite{cider}. Acc@50 is the accuracy of correctly retrieved inference sentences with 50 candidates to choose from. Unique is the number of inference sentences that are unique within the generated sentences, divided by the total number of sentences. Novel refers to the number of generated sentences that are not in the training data, divided by the total number of sentences. Text Given is when model is given any textual input during test time to generate the inferences. We bold the models based on the following order: 1) Best Text only model, 2) Best Image + Text model given visual and text input, 3) Best Image only model, and 4) Best Image + Text model given just visual input.} 
\label{tab:ablations}
\end {table}

\myparagraph{Automatic Evaluation}

Here, we describe the automatic evaluation measuring the quality of inference sentences. We first report the automatic metrics used in image captioning \cite{Chen2015MicrosoftCC}, such as BLEU-2 
\cite{bleu}, METEOR \cite{meteor}, and CIDER \cite{cider} across the 5 inferences.  Inspired by the metric in visual dialog \cite{Das2017VisualD}, we also use perplexity score to rank the ground truth inferences and inferences from the different image. We append negatives such that there are 50 candidates to choose from, rank each candidate using perplexity score, and  get the average accuracy of retrieved ground truth inferences (Acc@50 in Table \ref{tab:ablations}). Note that perplexity is not necessarily the perfect measure to rank the sentences, but good language models should still be able to filter out inferences that do not match the content in image and event at present. Lastly, we measure the diversity of sentences, so that we do not reward the model for being conservative and making the same predictions. We report the number of inference sentences that are unique within the generated sentences divided by the total number of sentences (Unique in Table \ref{tab:ablations}), and the number of generated sentences that are not in the training data divided by the total number of sentences (Novel in Table \ref{tab:ablations}). To capture the semantic diversity, we replace the predicted person tags with the same tag when calculating the above diversity scores.


\begin {table}[t!]
\begin{center}
\scalebox{0.8}{
\begin{tabular}{@{}l@{\ \ } c@{\ \ } c@{\ \ } c@{\ \ } | c@{\ \ } c@{\ \ }c@{\ \ }c@{\ \ }c@{\ \ }}
\toprule
\textbf{Modalities  } & \textbf{B-2} & \textbf{M  } & \textbf{C  } & \textbf{Human} & \textbf{Human} & \textbf{Human} & \textbf{Human} \\

\textbf{} & \textbf{ } & & & \textbf{  Before } & \textbf{  Intent  } & \textbf{  After  } & \textbf{  Avg  } \\

\midrule
\it{With Text Input.} \\
\midrule
Event + Place  
& 11.00 & 9.65 & 15.12 & 54.9 & 52.6 & 42.9 & 50.1  \\
Image + Event + Place + PG & \textbf{12.71} & \textbf{11.13 }& \textbf{17.36} &  \textbf{63.36} & \textbf{63.5} & \textbf{56.0} & \textbf{61.0} \\
\midrule
\it{Without Text Input.} \\
\midrule
No Input & 3.57 & 5.20 & 1.89 & 5.3 & 4.9 & 3.5 & 4.6 \\
Image + PG & 7.82 & 8.17 & 7.30 & 38.2 & 34.8 & 30.3 & 34.4\\ 
Image + Event + Place + PG + EP Loss & \textbf{9.33} & \textbf{10.12} & \textbf{10.82} & \textbf{42.9} & \textbf{36.8} & \textbf{34.8} & \textbf{38.2} \\

\midrule 
GT & - & - & - & 83.8 & 84.5 & 76.0 & 81.4 \\  
\bottomrule
\end{tabular}
}
\end{center}
\caption{\textbf{Generated Inference Results.} BLEU-2 (B-2) \cite{bleu}, METEOR (M) \cite{meteor}, CIDER (C), \cite{cider} and Human scores for the generated inferences on the Test split. We select 200 random images and generate 5 sentences for each of the three inference type (3000 sentences total). Then, we assign three annotators to determine if each inference sentence is correct, and take the majority vote. The models are chosen based on their best performance on the validation set when visual and/or textual modalities are available (bolded models in Table \ref{tab:ablations}).}
\label{tab:human}
\end {table}

\subsection{Results}

Table \ref{tab:ablations} shows our experimental results testing multiple training schemes and input modalities. We make the following observations: 1) Adding PG trick gives a boost for model over all metrics. 2) Model trained with both visual and textual (Image + Event + Place + PG) modalities outperform models trained with only one of modality (Event + Place; Image + PG) in every metric, including retrieval accuracy and diversity scores. This indicates that the task needs visual information to get higher quality inferences. 3) Adding place information helps in general. 4) Models with access to textual event and place information during test time, generate higher quality sentences than the same models without them (Text Given Yes vs No). This is not surprising as our dataset was collected with workers looking at the event, and the event already gives a strong signal understanding the content in the image. 
5) Lastly, adding the EP Loss boosts the performance if only the image content is available in the test time. This indicates that training the model to recognize events at present helps the performance, when the model has to generate inferences directly from image.

\myparagraph{Human Evaluation}

While the numbers in automatic evaluation give favorable results to our Image + Text model, they are not sufficient enough to evaluate the quality of generated inferences. We choose the best performing model when only image, text, or both inputs are available (model trained with no input and bolded models in Table \ref{tab:ablations}). We take 200 random images and the generated inferences, and ask the humans to evaluate their quality based on just the image content. Even for models that use ground truth inferences, we do not show the events to the workers and make them rely on image to make the decision. Specifically, we ask three different workers to evaluate if each inference is likely (1) or unlikely (0) to happen based on the image. We then take the majority out of three and calculate the average across all the inferences. 

Table \ref{tab:human} shows automatic metrics and human evaluation scores on the test split. We notice a similar pattern based on our automatic metric results: Image + Text model outperforms the Text only model (61.0 vs 50.1 on average) when text input is given in test time, and Image + Text model outperforms Image only model when text input is not given (38.2 vs 34.4 on average). We see that Text only model performs better than the Image + Text model without text input in test time, as the event sentence already describes the relevant details in the image and is a strong signal itself. Note that there is still a 20 point gap between our best model and ground truth inferences, meaning there is more room to improve our best model.


\subsection{Qualitative Examples}
Figure \ref{fig:qual1} presents some qualitative examples comparing the outputs of the various systems with the human annotated ground truth inferences. Overall, models that integrate information from both the visual and textual modalities generate more consistent and better contextualized predictions than models that only use either visual or textual information. 

Specifically, the first example (on the top) illustrates that in the absence of the event description, a model that solely relies on visual information generates incorrect predictions like ``order a drink at a bar", ``dance and have fun" etc. -- none of which are reasonable in the context of the event description. Similarly, a model that solely relies on the textual description, but not the visual information, generates ``get off of the stage" and even predicts ``her job as a scientist". This inference could be true in the absence of the visual features, but the image clearly shows that the person is in the audience, and not the one giving a presentation, nor she is portrayed as a scientist.

This pattern continues in the bottom example. [Person2] clearly looks worried but the Text only model predicts that he wants to ``alleviate his boredom", and does not incorporate this visual detail. Image only model again hallucinates wrong objects like ``have grabbed the wire". On the other hand, Image + Text model has the appropriate balance between the two models by stating there is possibly a criminal nearby as Person 2 is making an urgent call, and still predicts relevant visual details in the image.  Thus, we see that both visual and textual features contribute to generating coherent inferences. 

\begin{figure}[t]
\scriptsize
\begin{center}
\includegraphics[width=\linewidth]{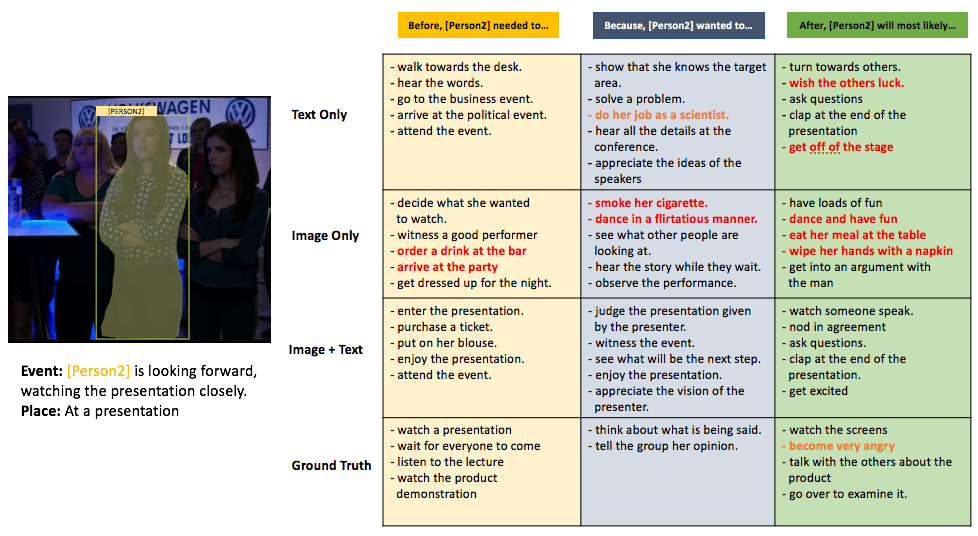} 
\includegraphics[width=\linewidth]{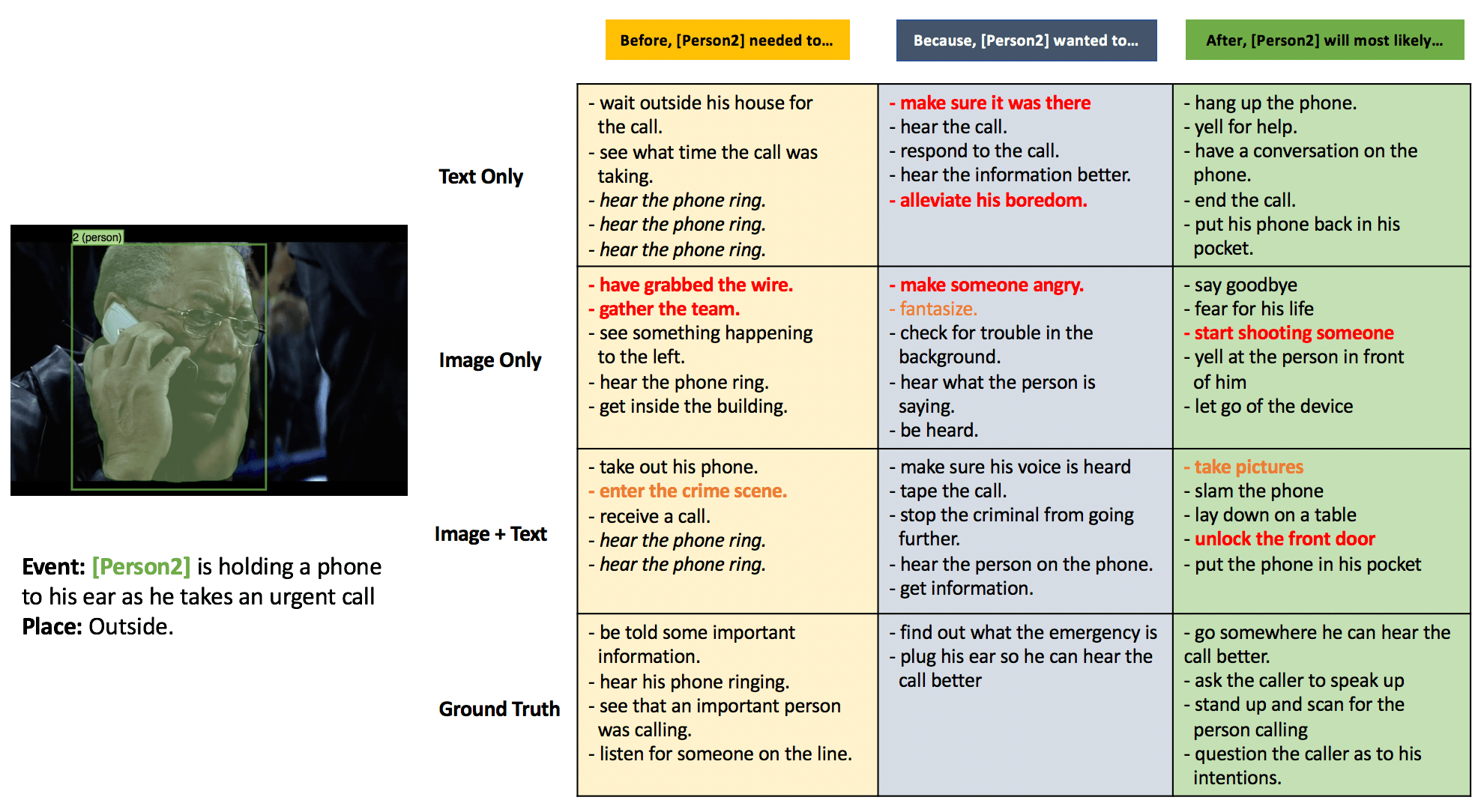}
\caption{\textbf{Qualitative Results.} Qualitative Examples comparing our best Text only, Image only, and Image + Text only model. Red highlights inference statements that are incorrect. Orange highlights if the sentences are plausible, but not expected. We see that our Image + Text model gives more consistent and contextualized predictions than the baseline models.}
\label{fig:qual1}
\end{center}
\end{figure}


\section{\label{sec:conclusion} Conclusion}
We present \papername, a novel framework of visual commonsense reasoning tasks to predict events that might have happened \textit{before}, events that that might happen \textit{after}, and the intents of people at \textit{present}. To support research in this direction, we introduce the first large-scale dataset of Visual Commonsense Graphs consisting of \numinf{} textual descriptions of visual commonsense inferences carefully annotated over a diverse set of 59,000 images. 

We present experiments with comprehensive baselines on this task, evaluating on two settings: 1) Generating inferences with textual input (event and place) and images, and 2) Directly generating inferences from images. For both setups, we show that integration between visual and textual commonsense reasoning is crucial to achieve the best performance.

\subsection*{Acknowledgements}

This research was supported in part by NSF (IIS1524371, IIS-1714566), DARPA under the CwC
program through the ARO (W911NF-15-1-0543),
DARPA under the MCS program through NIWC
Pacific (N66001-19-2-4031), and gifts from Allen Institute for Artificial Intelligence.

\clearpage
\section*{Supplemental Material}
\appendix
We provide detailed statistics about the \papername{}  dataset including its language diversity, and qualitative examples of inferences made by various model variants. We also show results from additional experiments for varying decoding schemes and performance for event description and place generation. 

Figure \ref{fig:v_comet_viz} shows a snapshot of our \textbf{Visual Commonsense Graphs}. The three images show very distinct scenes, but the graph allows us to reason that the \textit{intent} of the person sitting at a shack (bottom right image), the \textit{before} event for the woman at an indoor bar (top left image), and the likely \textit{after} event for the woman in the ballroom (bottom left) are identical -- to ``order a drink". Each image is associated with several inferences of the three types: (i) intents at present, (ii) events before, and (iii) events after.

\begin{figure}[ht!]%
    \centering
    \includegraphics[width=30pc]{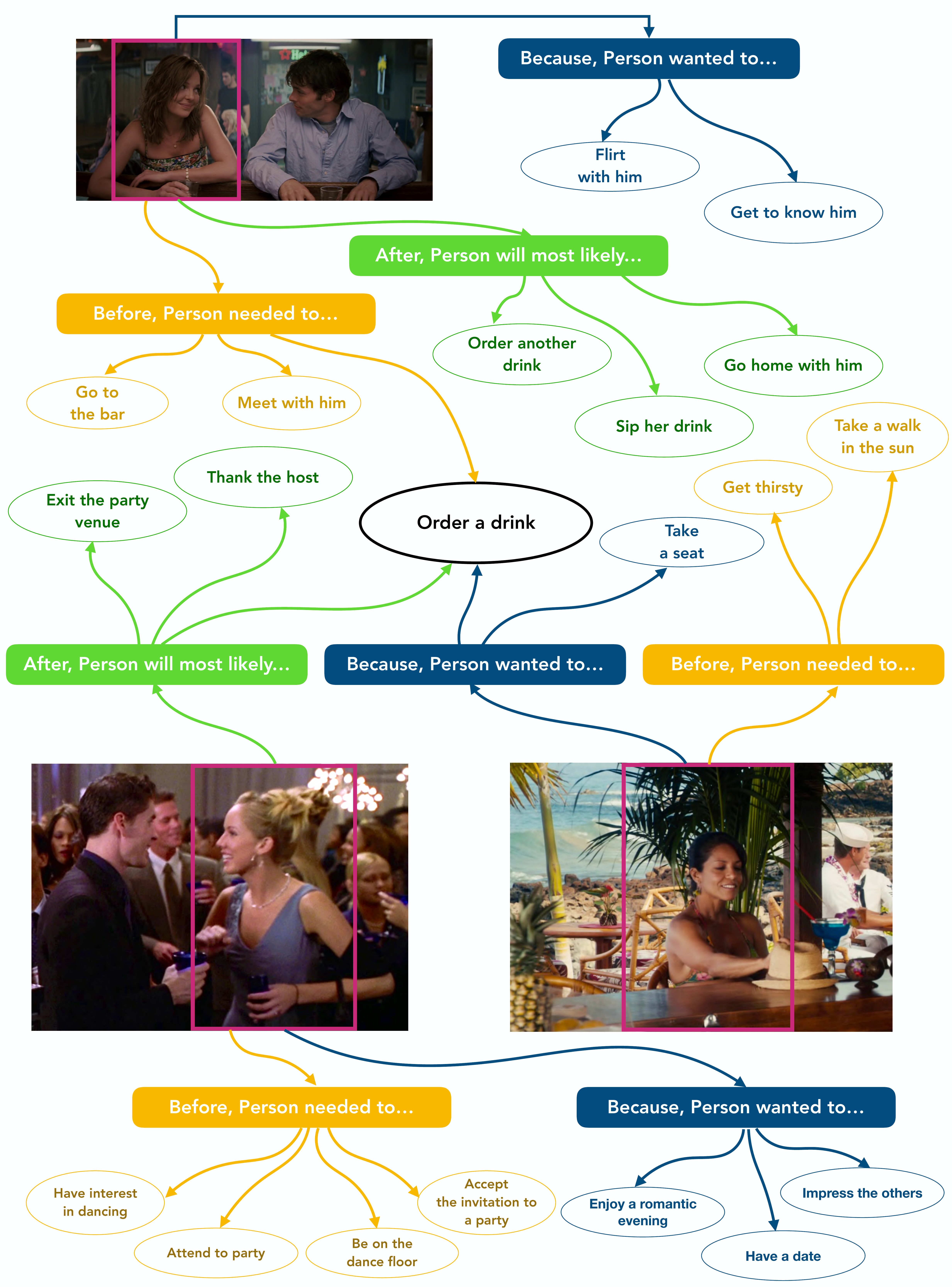} \\ 
    \caption{Snapshot of our \textbf{Visual Commonsense Graphs}. Images from very distinct scenes are connected by the same inference sentence ``order a drink".}
    \label{fig:v_comet_viz}
\end{figure}

\section{Dataset Statistics}
\begin{figure}[ht!]%
    \centering
    \includegraphics[width=\textwidth]{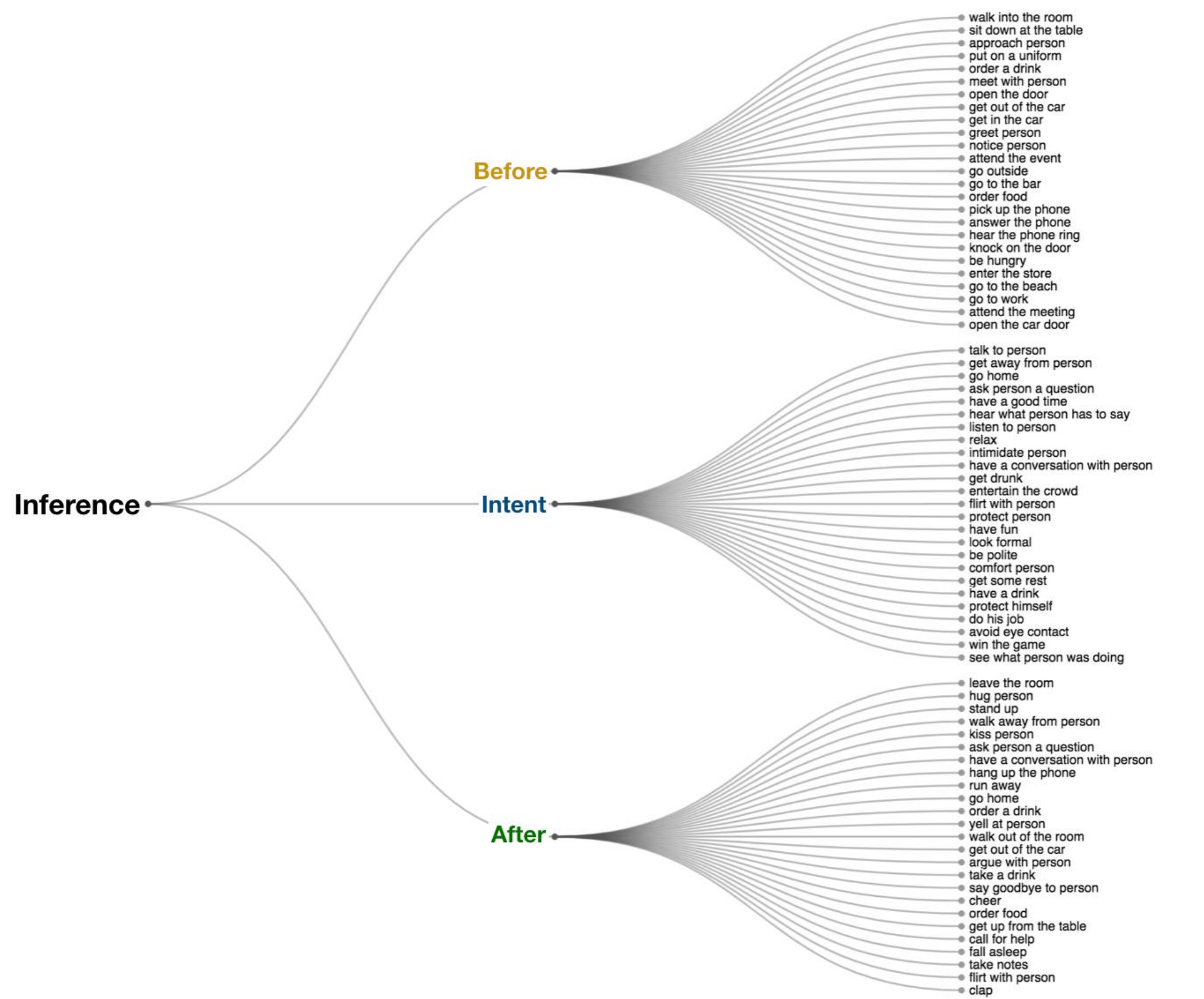} \\ 
    \caption{Most frequent phrases mentioned per Inference Type}
    \vspace{1cm}
    \label{fig:inf_viz}
\end{figure}

Additional statistics of the dataset are provided in Table \ref{tab:statistics}. On average, there are 2.12 \textit{Intent}, 4.30 \textit{Before}, and 4.31 \textit{After} Inferences for each event. Each image has 2.34 events on average (place is always annotated once for each image). Figure \ref{fig:inf_viz} shows a breakdown of most frequent phrases per each inference type. \textit{Before} and \textit{After} inferences tend to focus on action statements, specifically activities involving entering or leaving the place.  \textit{Intent} inferences mostly involve various interactions with another person and also include person's mental states, such as ``have a good time", ``be polite", and ``look formal".


We also provide more detailed distribution of the sentences. Figure \ref{fig:bigram} shows the number of occurrences of starting bigram (first two words) for each inference type. As we see, the distribution is vastly different based on the inference type, and there is no overlapping bigram among the top 5 phrases. Figure \ref{fig:event} shows the a) noun and b) verb distributions of the event sentences. We omit person in noun, and linking verbs in verb distributions for visualization purposes. We show histogram of unique place phrases in Figure \ref{fig:place}. Popular places that are annotated include ``office", ``living room", ``restaurant", ``kitchen", and ``party". Lastly, Figure \ref{fig:length} provides the length of event, place, and inference sentences.

\begin {table}[t]
\begin{center}
\setlength\tabcolsep{0pt} 
\begin{tabular}{@{}l@{}c@{}@{}c@{}@{}c@{\ \ }}
\toprule
\textbf{} & \textbf{       } & \textbf{Avg Count}\\
\midrule 
\# of \textit{Intent} Inference per Event & & 2.12 \\
\# of \textit{Before} Inference per Event & & 4.30 \\
\# of \textit{After} Inference per Event & & 4.31 \\
\# of Event per Image & & 2.34\\ 
\midrule
\# of Unique Persons Mentioned in Event & & 1.51 \\ 
\# of Unique Persons Mentioned in Inference & & 0.27 \\
\midrule
\# of Words in Event & & 9.93\\
\# of Words in Place& & 3.44 \\
\# of Words in Inference& & 4.8\\
\bottomrule
\end{tabular}
\end{center}
\caption{Additional Statistics for \papername{}.}
\label{tab:statistics}
\end{table}


\section{Qualitative Examples}
We show more qualitative examples in Figure \ref{fig:qual1} and \ref{fig:qual2}. Following Figure 6 of the main paper, we use the best performing model when Text only, Image only, and Image + Text input are given. Specifically, the models are Row 3 [Event + Place], Last Row [Image + Event + Place + PG + EP Loss (No Text Given)], and Row 8 [Image + Event + Place + PG] in Table 2 of the main paper. We highlight obviously incorrect inference sentences as red, and plausible but not expected as orange.  

Figure \ref{fig:qual1}(a) shows Person1 [P1] serving food and ``putting a platter on the table". While the event and place information does not mention that [P1] is a waiter, our Image + Text model uses the visual information to correctly infer that he needed to ``be hired as a waiter at a formal event". The model also generates inferences that involve other relevant people (e.g. ``serve [P2], [P4], [P5]"). Text only model fails to infer that [P1] is a waiter and sees him as the one joining the meal. For example, the model generates ``ask [P2] for a menu" and ``sip the water" in the \textit{after} inferences. Image only model can generate inferences involving other people and recognize that the place is a restaurant; however, it fails to get the detail that [P1] is the one serving the food. Figure \ref{fig:qual1}(b) shows an example focusing on person's mental state. While the image takes place at an outdoor party, it is unlikely that Person2 [P2] will dance, based on the event ``is alone and feeling awkward" and her passive body language. We see that Image only and Text only models fail to incorporate this information and generate typical activities at a party, such as ``dancing" or ``drinking". Image + Text model makes inferences that suggests [P2] is not having fun and even predicts that she might ``return to her car and drive away" or ``yell at the people" after the event. Additional examples are shown in Figure \ref{fig:qual2} and we see that Image + Text model generates more coherent and plausible inferences. 

\myparagraph{Inference vs Captioning}
Figure \ref{fig:diff_task} shows an example highlighting the main difference between our task and other visual captioning models. For fair comparison with image captioning models, we show the inference sentences using Image only model in Figure \ref{fig:diff_task} (a). Top of Figure \ref{fig:diff_task} (b) shows results from dense captioning model \cite{Johnson2015DenseCapFC} that predicts the bounding boxes and associated captions. Bottom of the figure provides five captioning outputs using the strong baseline in \cite{anderson2018bottom}. We see that captioning models are mostly correct, such as the phrase ``A woman is wearing a black shirt" and caption ``a group of people sitting around a laptop". The descriptions, however, miss the detail of people working in the office. On the other hand, our Image only model can go beyond the simple details of sitting down at the desk and generate more contextualized information in office environment, such as ``arrive at work early to get an interview", ``see what was on the computer", and ``gather up all her files". Using our visual commonsense graphs, we see that we can infer more salient and detailed information in still images that captioning tasks fail to provide.

\begin{figure}[ht!]
\scriptsize
\begin{center}
\subfloat[]{\includegraphics[width=\textwidth]{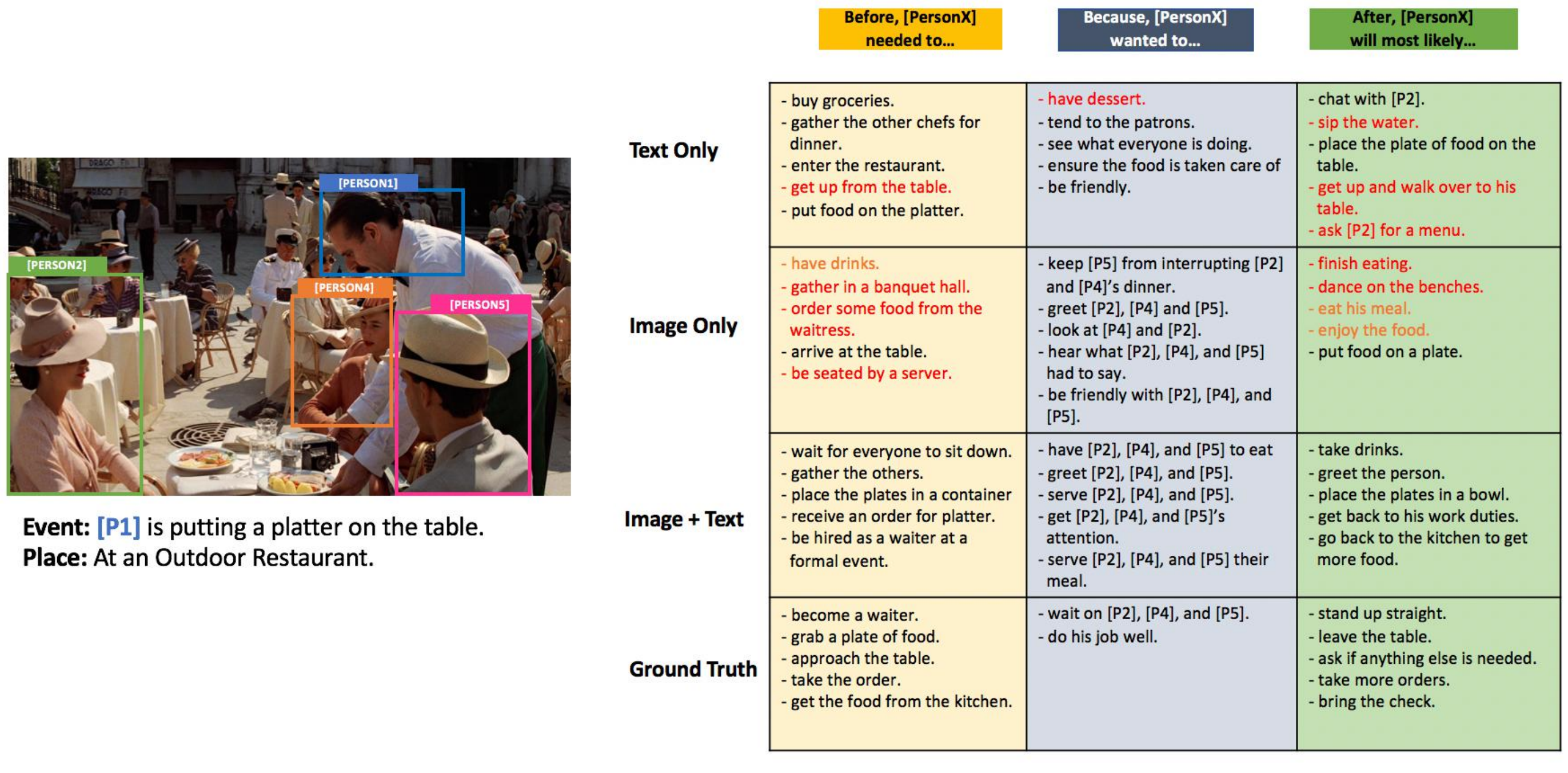}}  \\
\subfloat[]{\includegraphics[width=\textwidth]{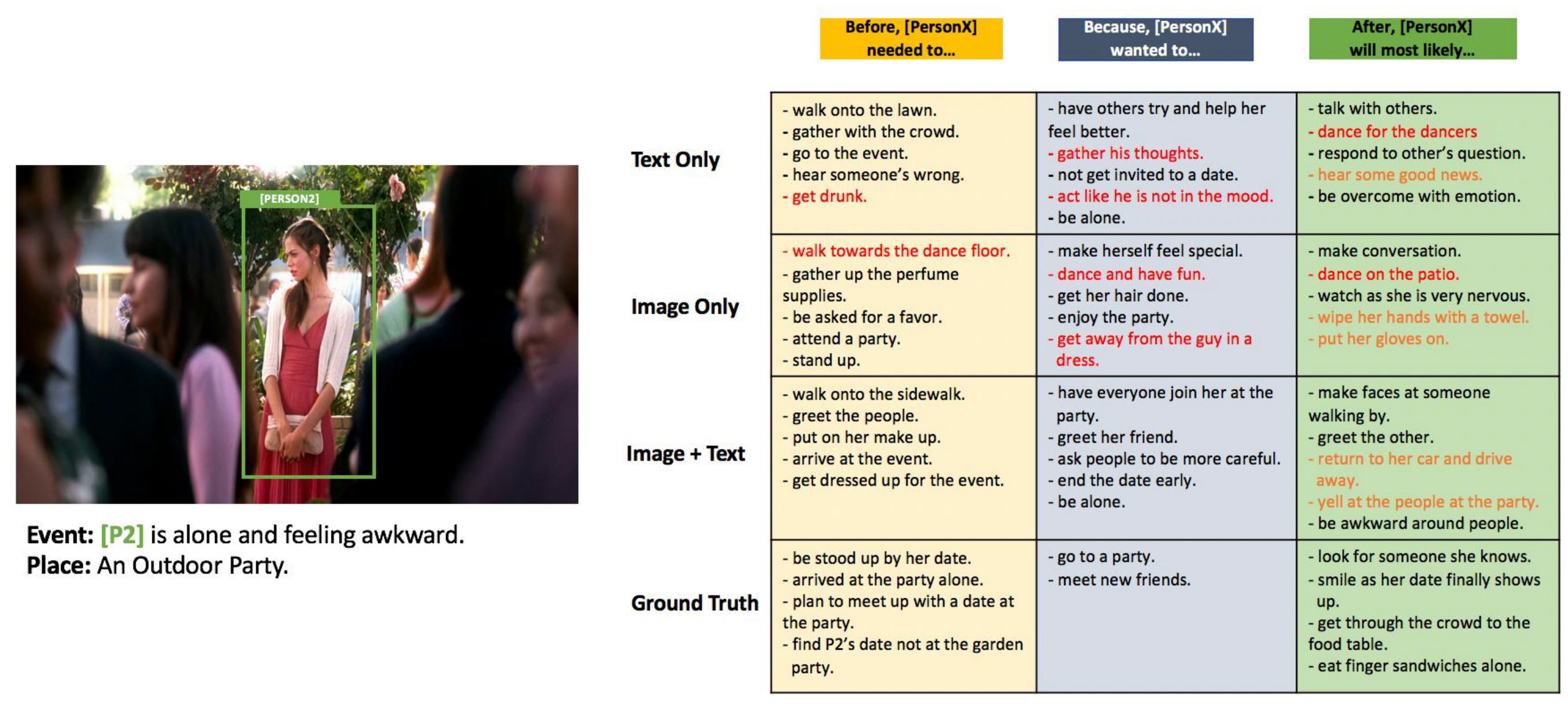}}
\caption{\textbf{Qualitative Results.} Qualitative Examples comparing our best Text only, Image only, and Image + Text model. Red highlights inference statements that are incorrect. Orange highlights if the sentences are plausible, but not expected. [PersonX] in the inference type refers to the subject of the event.}
\label{fig:qual1}
\end{center}
\end{figure}

\begin{figure}[ht!]
\scriptsize
\begin{center}
\subfloat[]{\includegraphics[width=\textwidth]{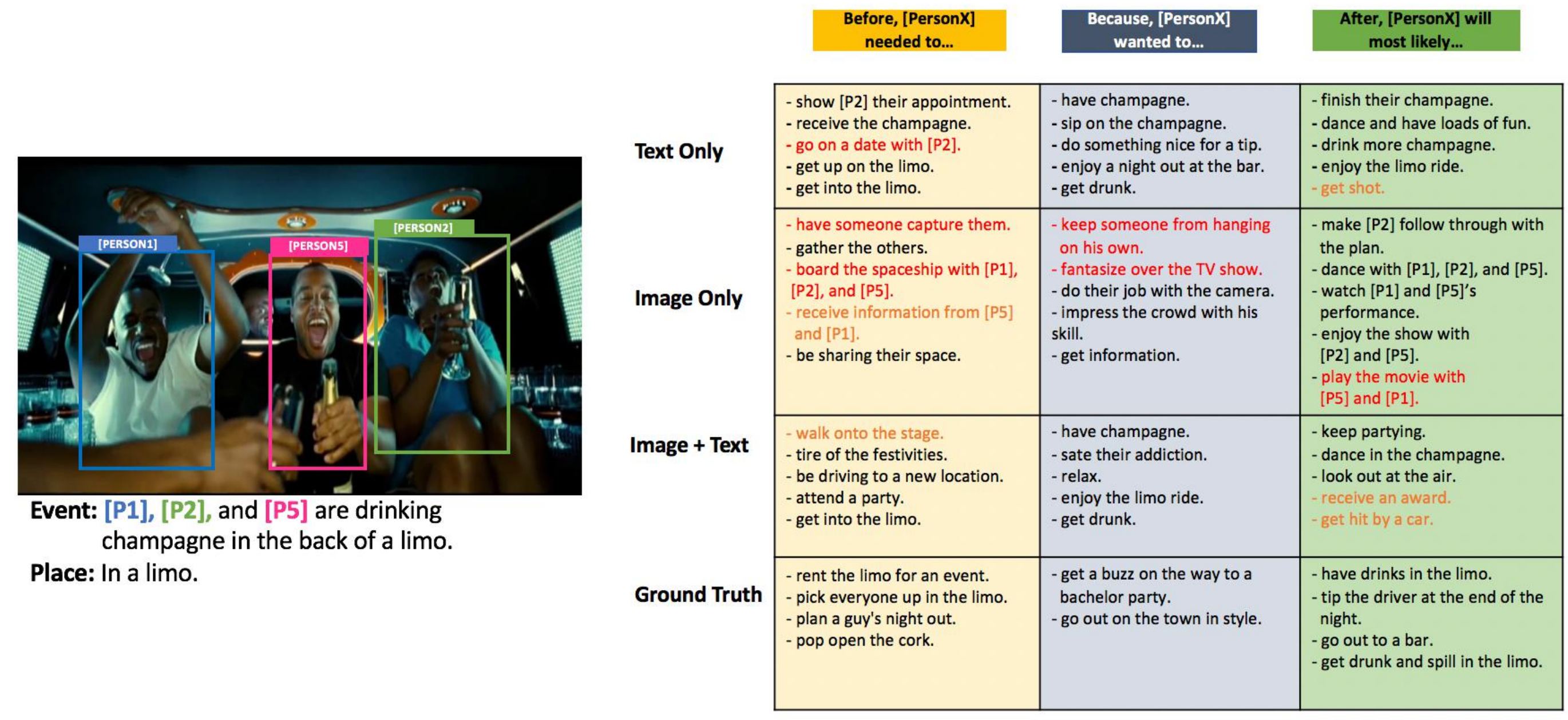}}  \\
\subfloat[]{\includegraphics[width=\textwidth]{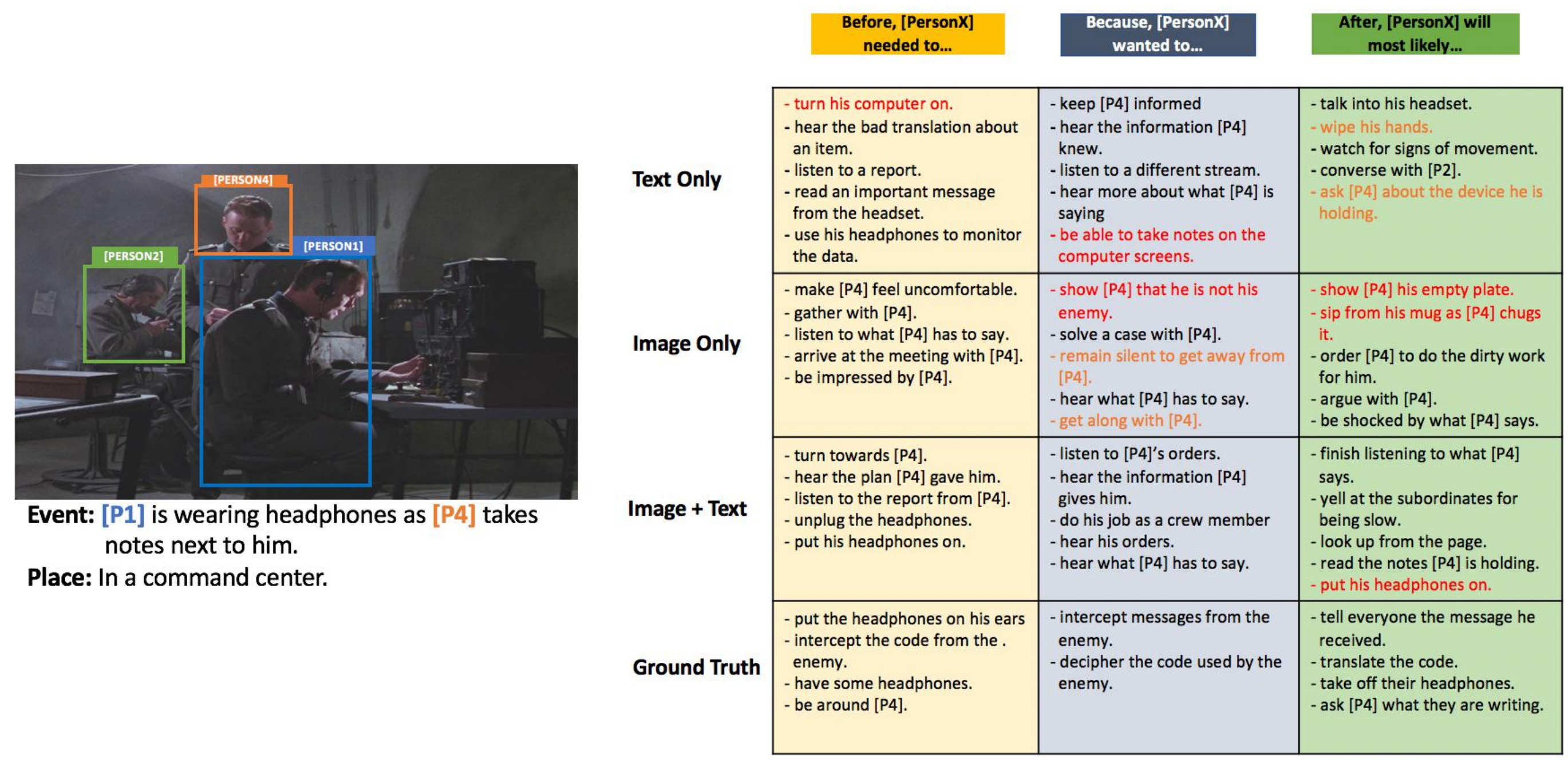}}
\caption{\textbf{Qualitative Results.} Qualitative Examples comparing our best Text only, Image only, and Image + Text model. Red highlights inference statements that are incorrect. Orange highlights if the sentences are plausible, but not expected. [PersonX] in the inference type refers to the subject of the event.}
\label{fig:qual2}
\end{center}
\end{figure}

\begin{figure}[ht!]
\scriptsize
\begin{center}
\subfloat[Inference with Image Only Model (event and place are not taken as input, and shown just for visualization)]{\includegraphics[height=0.4\textheight]{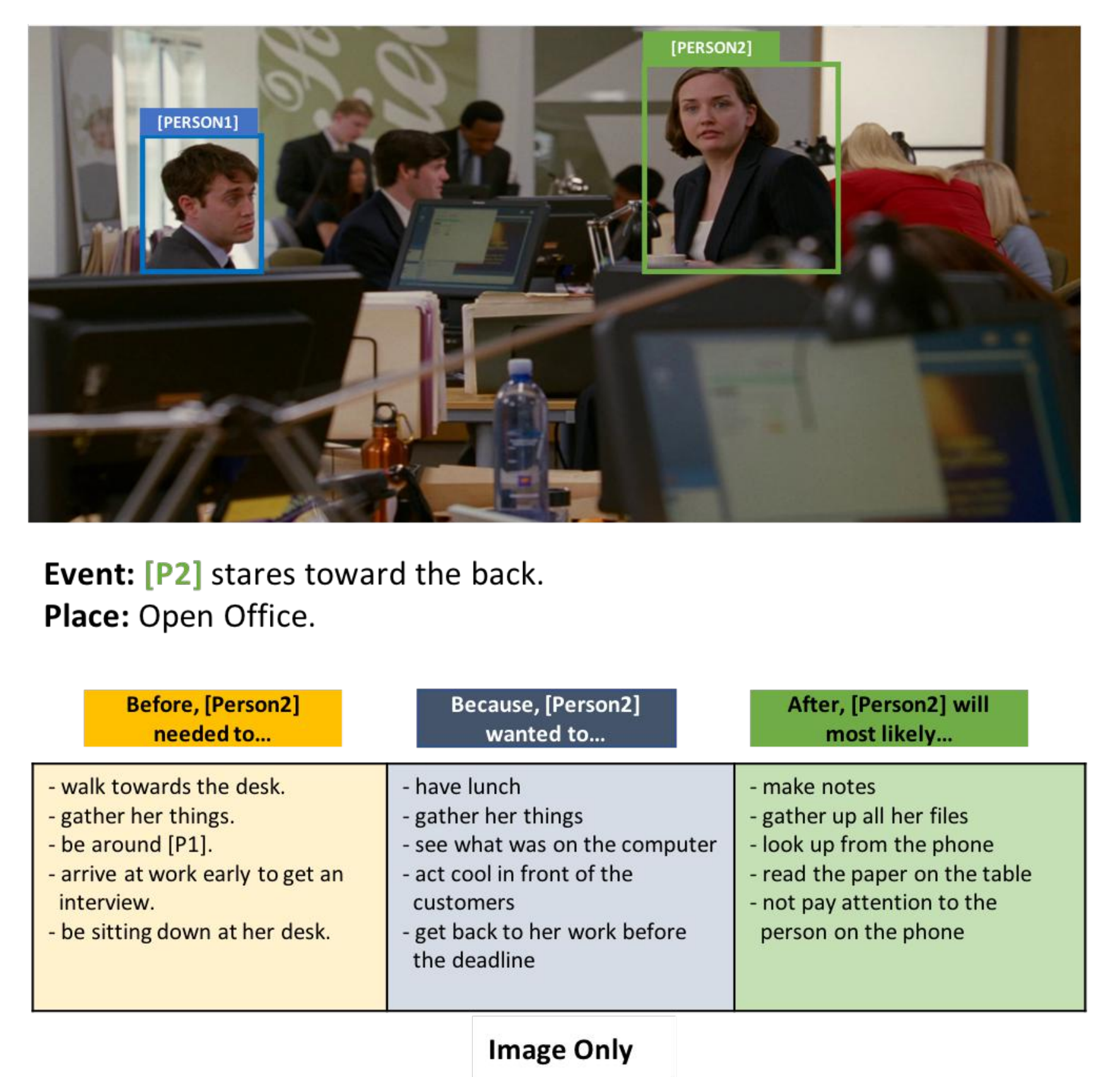}}  \\
\subfloat[Results from Dense Captioning \cite{Johnson2015DenseCapFC} and Bottom-up and Top-down image captioning model \cite{anderson2018bottom}]{\includegraphics[height=0.4\textheight]{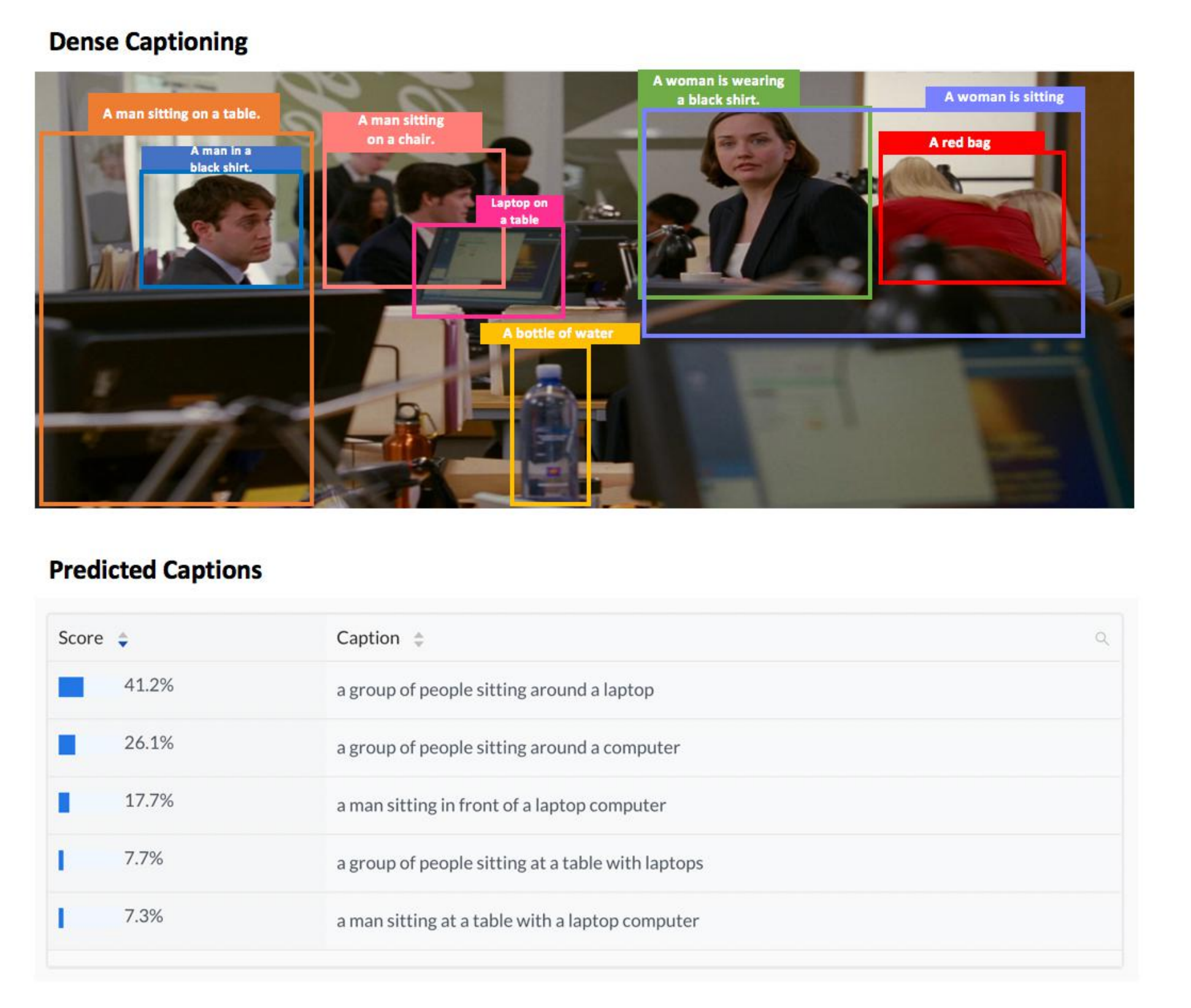}}
\caption{\textbf{Difference between Inference and Captioning}. We see that our task (a) generates sentences that are more diverse and rich in content than the captioning models (b).}
\label{fig:diff_task}
\end{center}
\end{figure}

\section{Annotation Template}
\begin{figure}[t]
\scriptsize
\begin{center}
\subfloat[We annotate event, place, and intent inferences in the First Annotation Stage.]{\includegraphics[height=0.43\textheight]{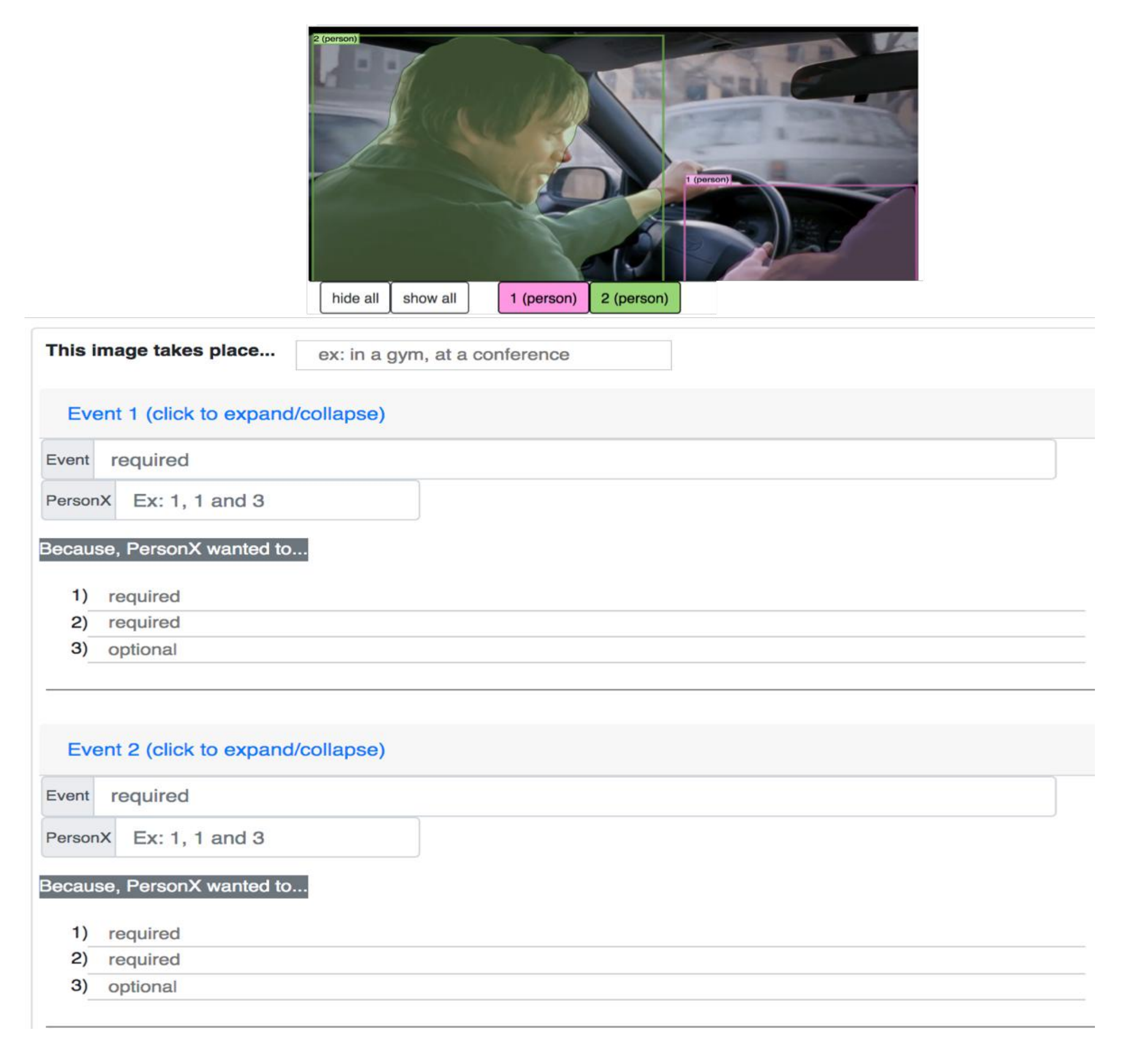}}  \\
\subfloat[We annotate before and after inferences in the Second Annotation Stage.]{\includegraphics[height=0.4\textheight]{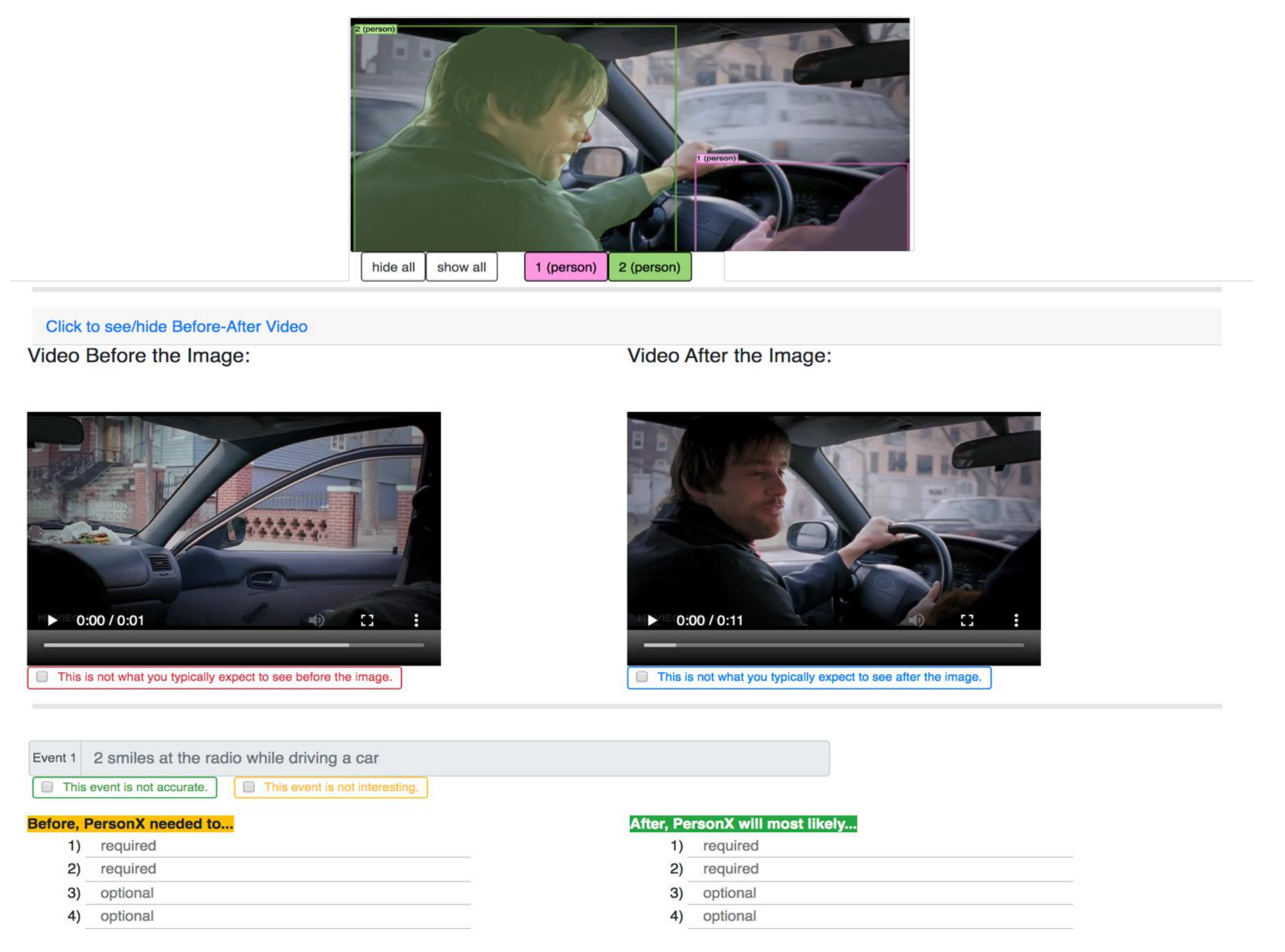}}
\caption{\textbf{Our Two-Stage Annotation Pipeline}. See Section C for more details.}
\label{fig:annot}
\end{center}
\end{figure}

Figure \ref{fig:annot} shows the template used for our two stage annotation pipeline. The first stage Figure \ref{fig:annot}(a) involves writing at least two events and place per image. Then, each event is given optional choice of writing 2-3 \textit{intent} inferences. Note only one worker is assigned for each image in the first stage. In the second stage Figure \ref{fig:annot}(b), each event is then annotated with 2-4 \textit{before} and \textit{after} inferences. Here, we assign two distinct workers to get the two inferences. In sum, each event is annotated with at least 10 inference sentences. 

\section{Decoding Strategies}
In the main paper, the inference sentences are generated using Nucleus Sampling \cite{Holtzman2019TheCC}, which is the state of the art decoding method to get more coherent and diverse sentences. Another option is to use beam search, which has shown to perform well in language metric but provides far less diverse sentences \cite{Vijayakumar2016DiverseBS}. This is especially problematic for generating \textit{multiple} inferences, where we want to avoid generating duplicating phrases within the inference set. 

Table \ref{tab:decoding} shows the comparison between the two decoding schemes and generate 5 sentences for each inference. We use the models from Row 3, 8, 10, and 12 in Table 2 of the main paper. We report BLEU-2 \cite{bleu}, and diversity metrics, such as proportion of unique inferences (UI), and ratio of unique unigrams/bigrams to number of words within the set of 5 sentences (DIV1/2-S) \cite{shetty2017speaking}. In language metric, we see that the model performance is consistent regardless of the decoding strategy: Image + Text model (Image + Event + Place + PG) outperforms other Text only and Image only baselines for Nucleus Sampling and beam search. Image + Text model also gets the most number of unique sentences for the both decoding schemes. While BLEU-2 \cite{bleu} scores are higher using beam search, we see that the diversity scores are much worse. Specifically, UI drops by half, and DIV1/2-S scores also suffer for the best performing model. We also see that Nucleus Sampling gets similar DIV1/2-S to the ground truth across all models, while there is around 30 and 20 point gap respectively for beam search methods. Note that getting the highest DIV1/2-S does not necessarily indicate having the highest diversity if these scores above a certain threshold. For instance, the model trained with No Input gets the highest DIV1-S and even higher than ground truth sentences, while UI is close to 0.

\begin{figure}[ht!]%
    \centering
    \includegraphics[height=0.4\textheight]{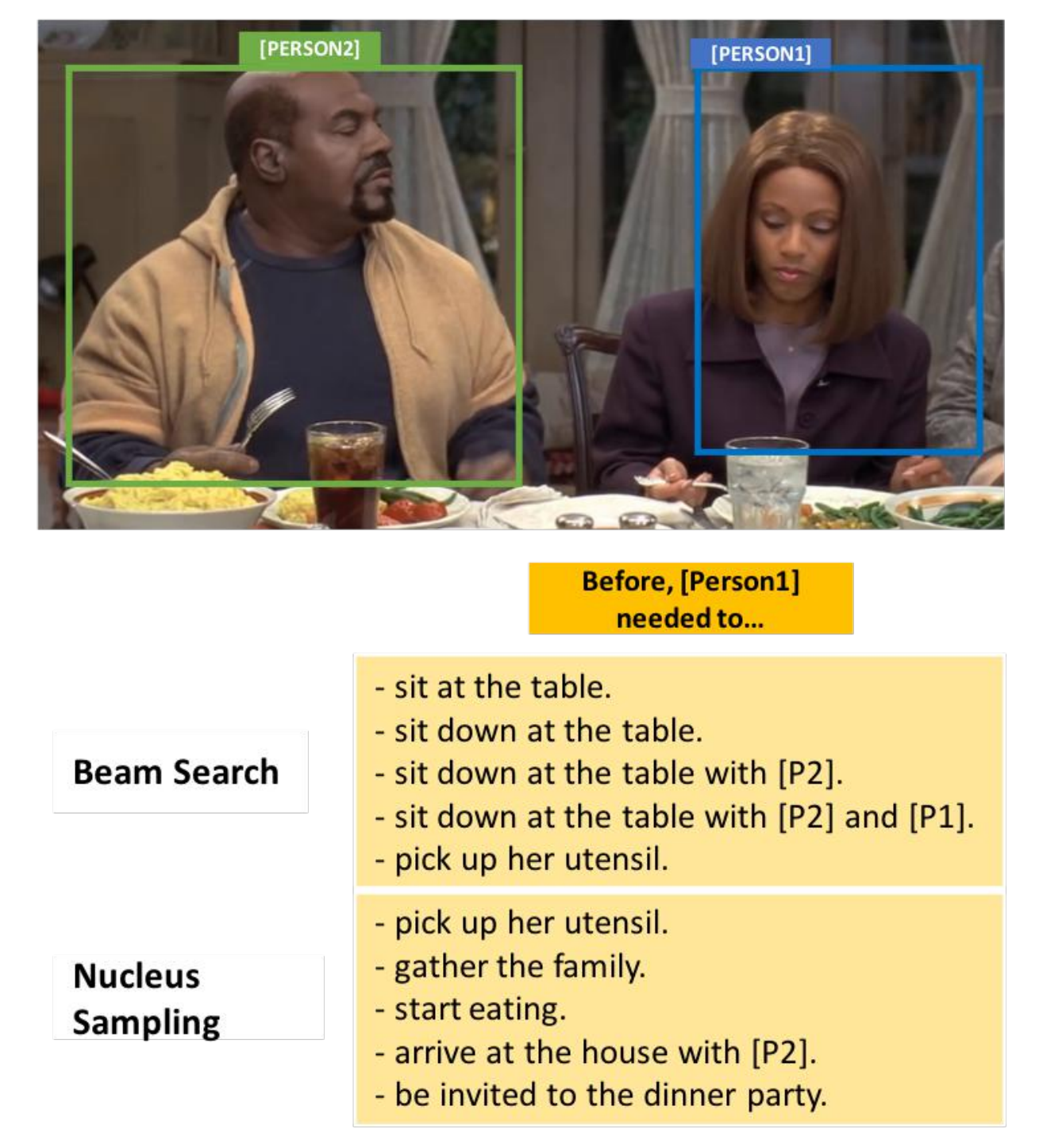} \\ 
    \caption{Comparison between beam search and Nucleus Sampling from the same model. We see that beam search repeats the phrase ``sit down at the table", while Nucleus Sampling gets more diverse and richer sentences.}
    \label{fig:bs}
\end{figure}

\begin{figure}[t]%
    \centering
    \subfloat[Before]{{\includegraphics[width=\linewidth, height=0.29\textheight]{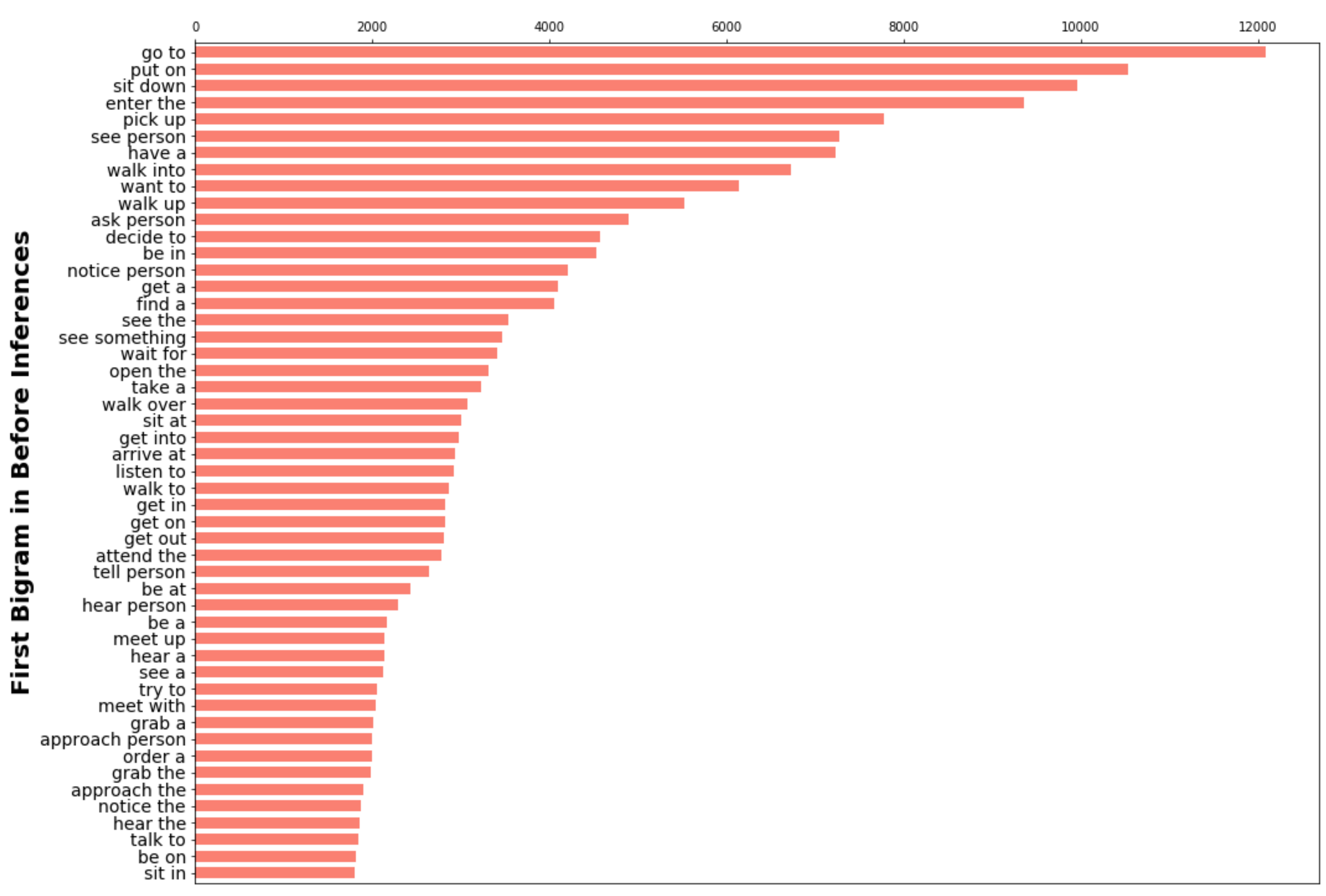} }}
    \\
    \subfloat[Intent]{{\includegraphics[width=\linewidth, height=0.29\textheight]{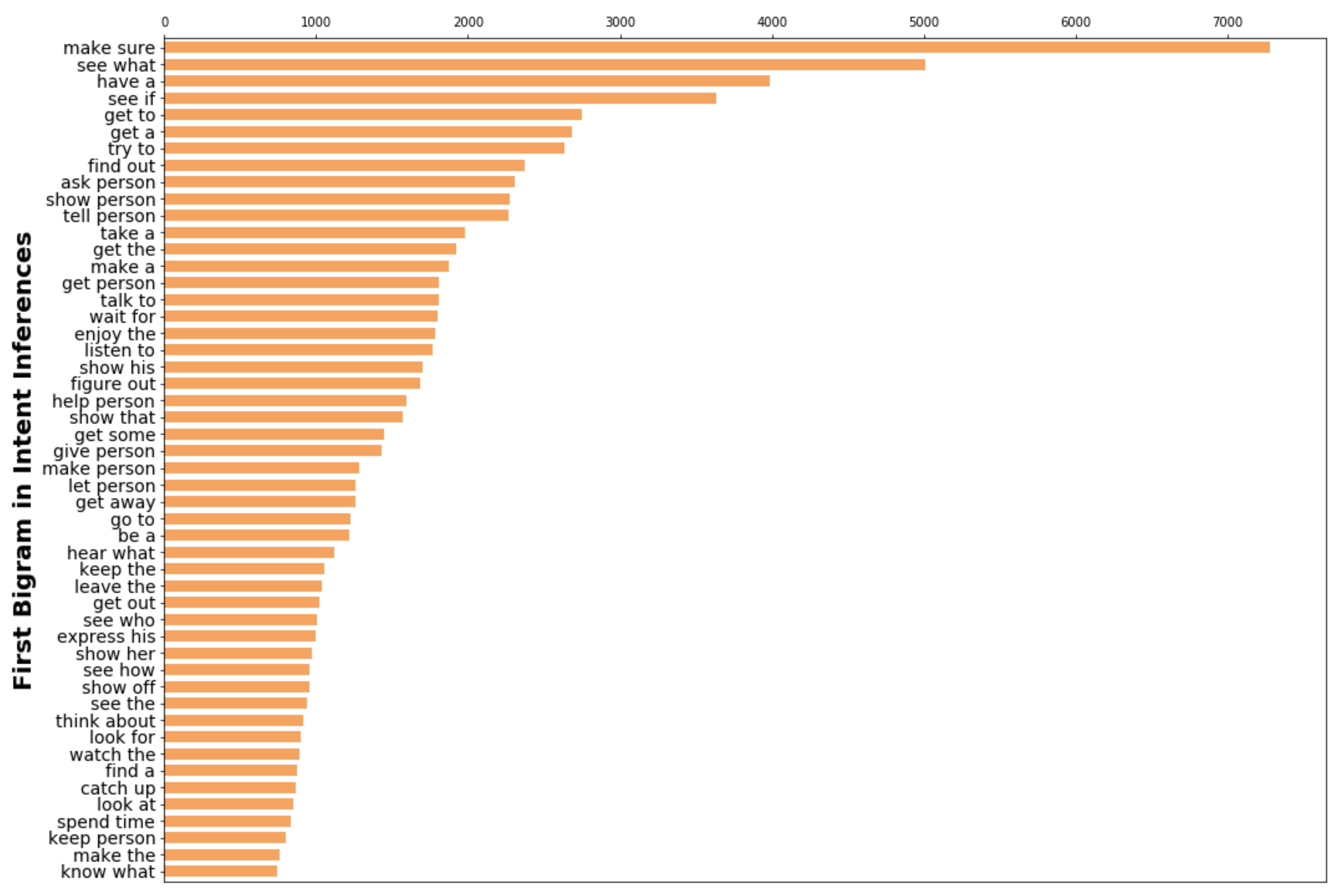} }}
    \\
    \subfloat[After]{{\includegraphics[width=\linewidth, height=0.29\textheight]{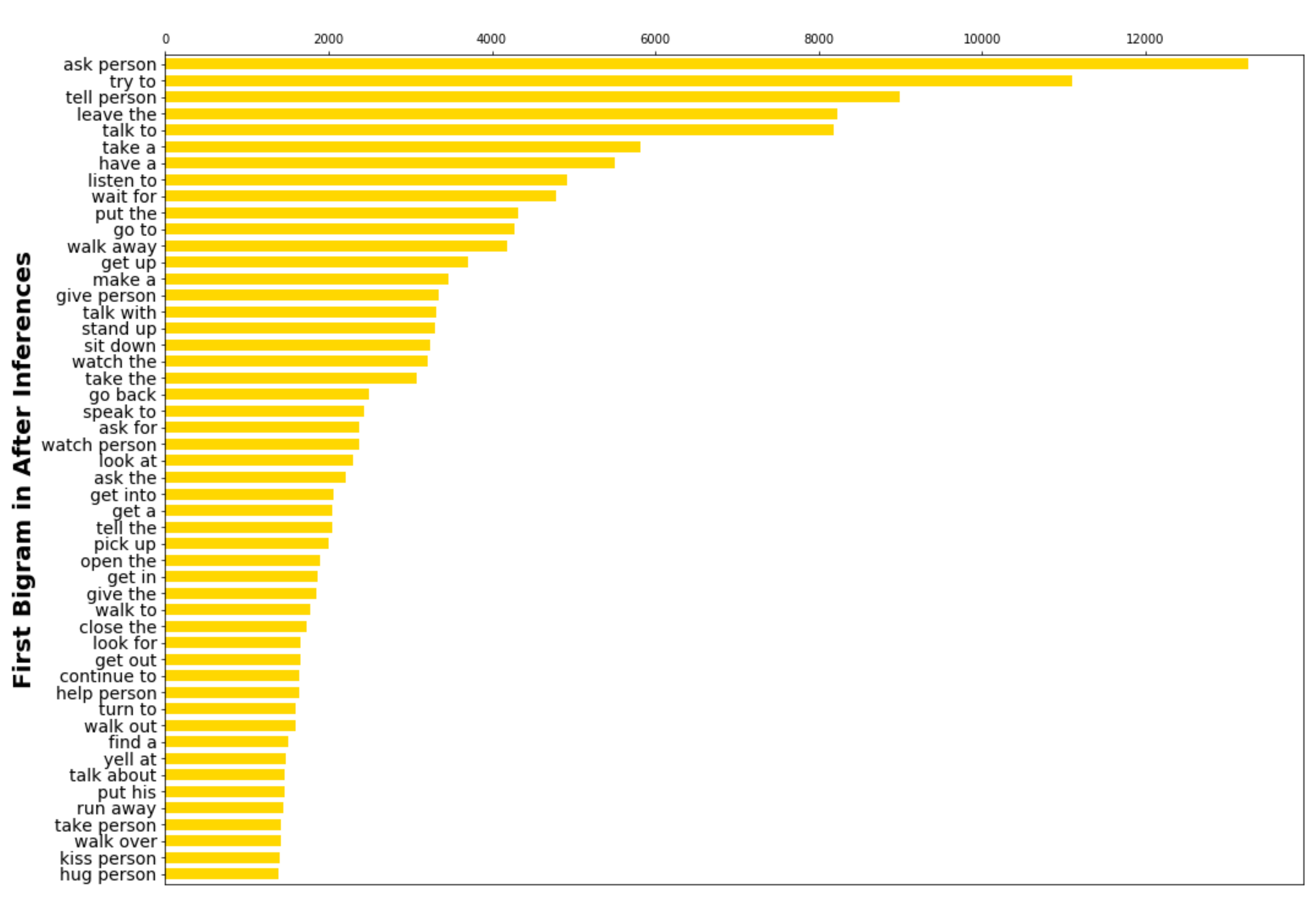} }}
    \caption{Most Frequent Starting bigram in a) Before, b) Intent, and c) After Inferences.}
    \label{fig:bigram}
\end{figure}

\begin{figure}[t]%
    \centering
    \subfloat[Nouns in Event Sentences]{{\includegraphics[width=\textwidth, height=0.45\textheight]{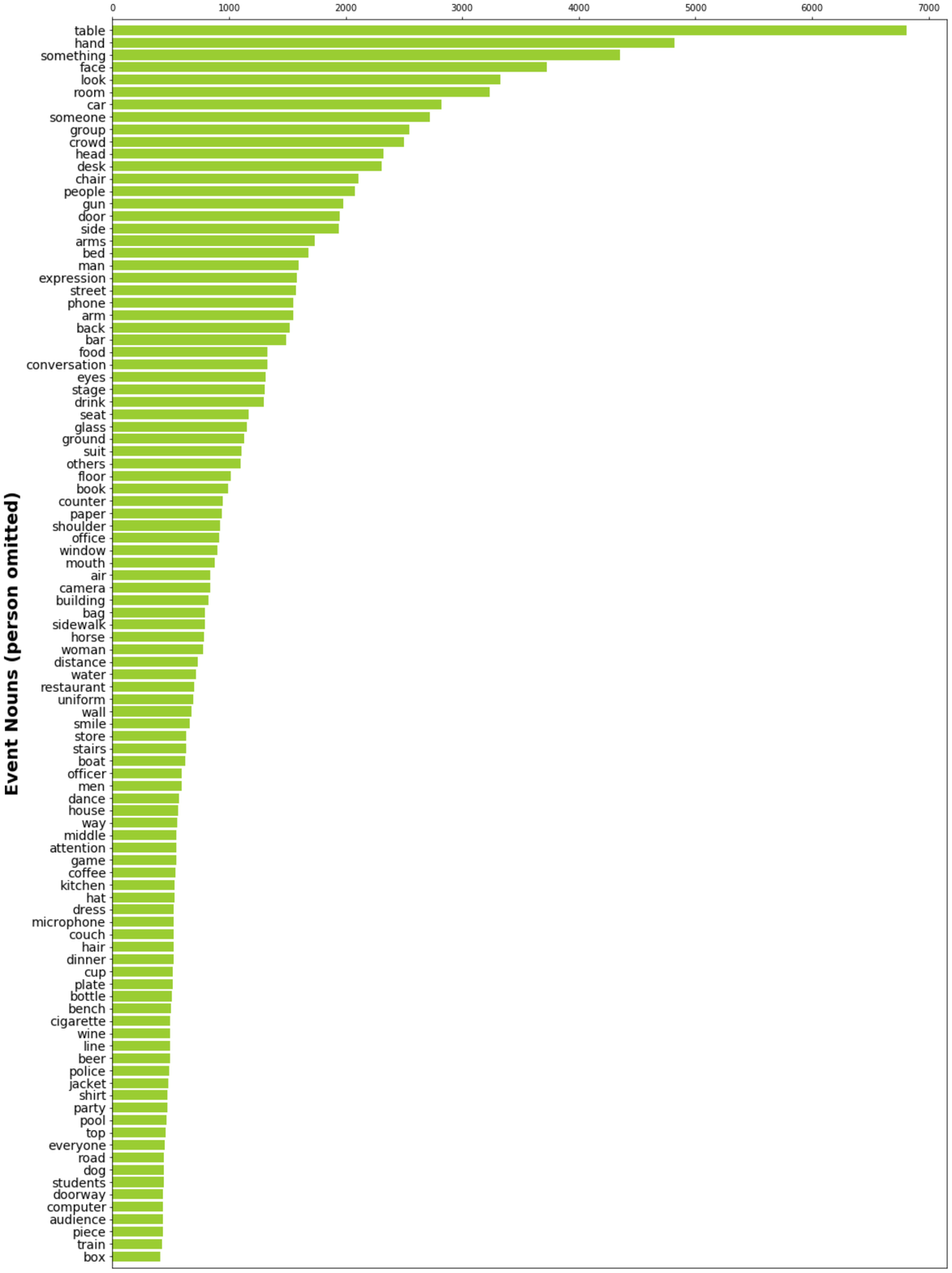} }}
    \qquad
    \subfloat[Verb Phrases in Event Sentences ]{{\includegraphics[width=\textwidth, height=0.45\textheight]{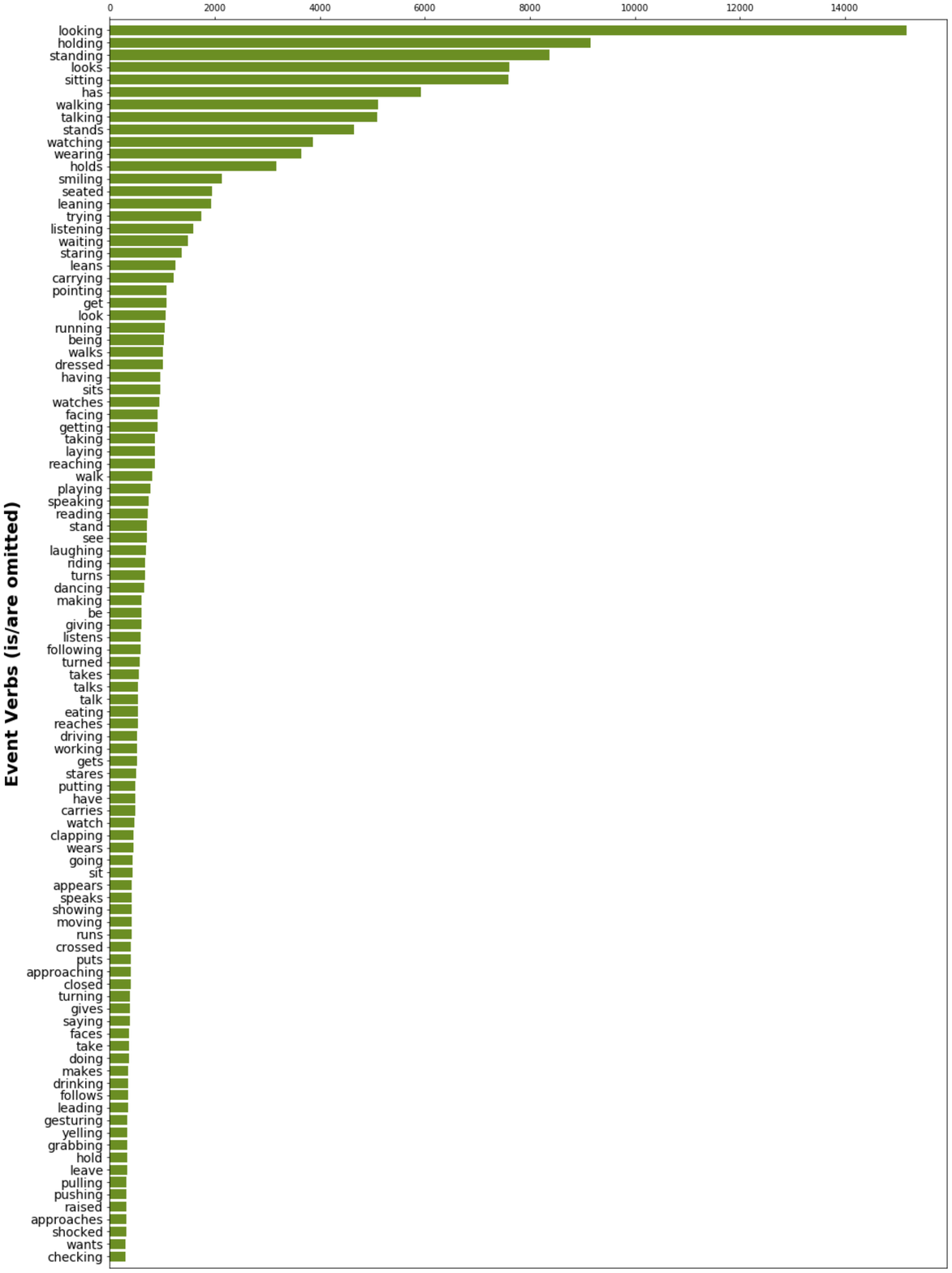} }}
    \caption{Most Frequent Noun \& Verbs in Event Sentences}
    \label{fig:event}
\end{figure}

\begin{figure}[t]%
    \centering
    \includegraphics[width=\textwidth, height=0.9\textheight]{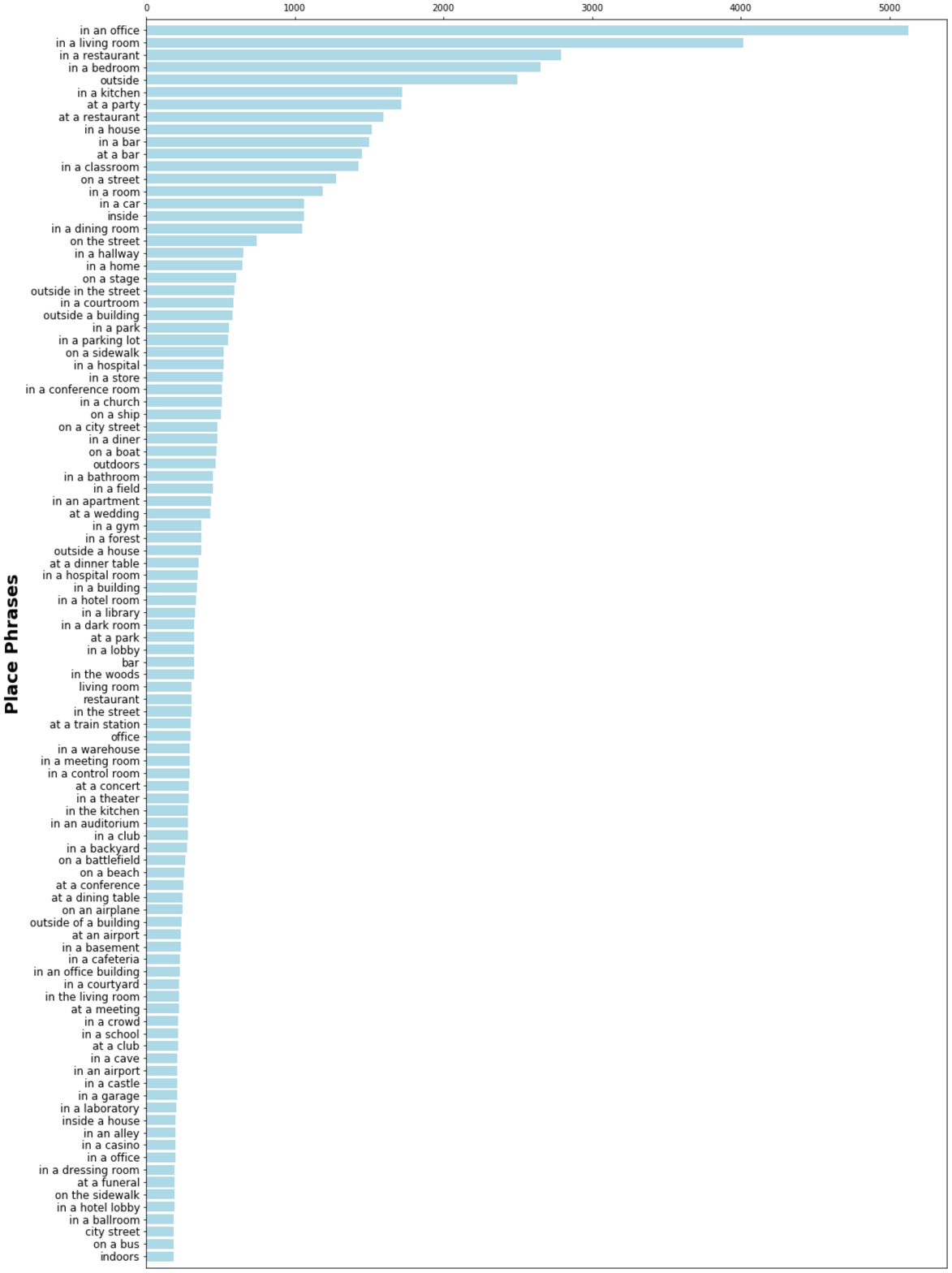}
    \caption{Place Phrases}
    \label{fig:place}
\end{figure}

\begin{figure}[t]%
    \centering
    \subfloat[Number of Words in Event]{{\includegraphics[width=0.9\textwidth]{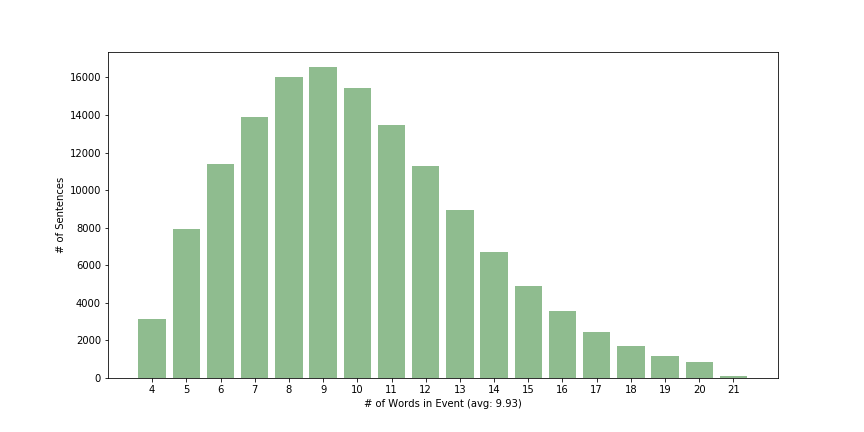} }} \\ 
    \subfloat[Number of Words in Place]{{\includegraphics[width=0.9\textwidth]{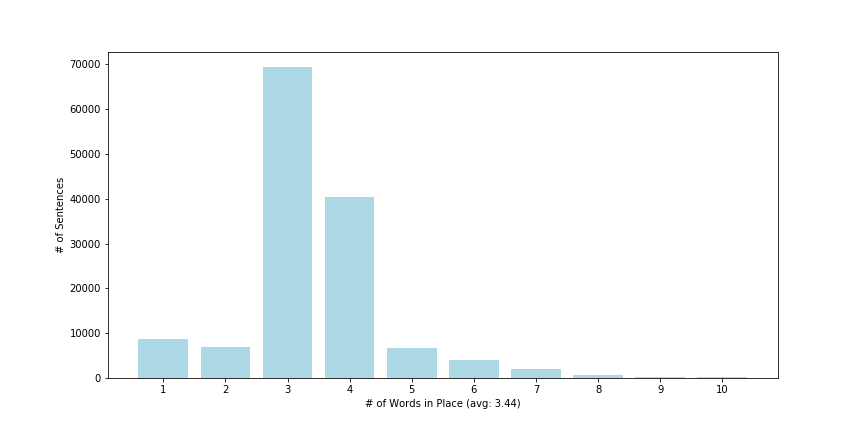} }} \\ 
    \subfloat[Number of Words in Inference]{{\includegraphics[width=0.9\textwidth]{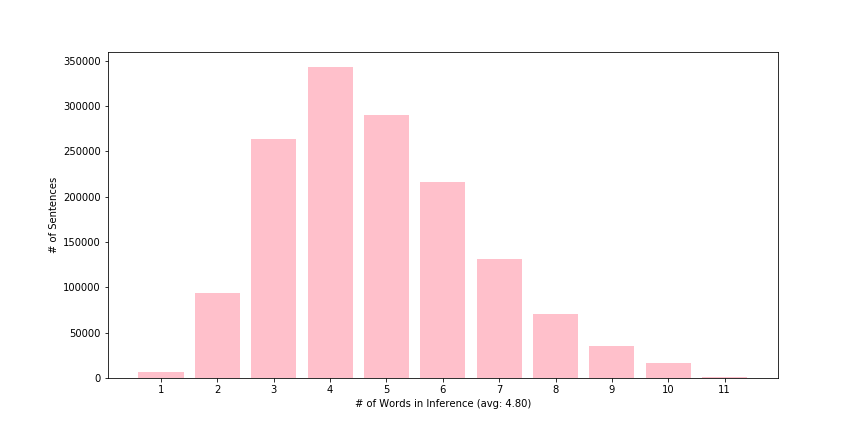} }}
    \caption{Sentence Length}
    \label{fig:length}
\end{figure}

Figure \ref{fig:bs} qualitatively shows the problem of using beam search over sampling methods. Beam search is prone to repeating the same phrases across the set, such as ``sit down at the table", which are correct but not desirable for our task. On the other hand, Nucleus Sampling captures correct inference statements but also diverse and rich in content. This suggests that sampling based decoding scheme is far preferable to beam search, when generating multiple candidates.

\begin {table}[ht!]
\begin{center}
\setlength\tabcolsep{0pt} 
\begin{tabular}{@{}l@{}c@{}@{}c@{}@{}c@{}@{}c@{}@{}c@{}@{}c@{}@{}l@{}}
\toprule
\textbf{Modalities} & \textbf{  BLEU-2 $\uparrow$ } & \textbf{  UI$\uparrow$  } & \textbf{  DIV1-S  } & \textbf{  DIV2-S  }\\
\midrule 
\it{Nucleus Sampling} \\
\midrule
No Input & 4.88 & 0.00 & 89.30 & 75.20\\
Event + Place & 10.49 & 47.42 & 82.89 & 75.22 \\
Image + PG. & 7.84 & 35.62 & 83.70 & 75.99\\
Image + Event + Place + PG. & \textbf{11.76} &  \textbf{51.99} & 80.36 & 74.89 \\
\midrule
\it{Beam Search} \\
\midrule
No Input & 7.36 & 0.00 & 54.00 & 48.70 \\
Event + Place & 18.97 & 23.64 & 56.10 & 54.50 \\
Image + PG. & 13.21 & 8.79 & 53.91 & 52.75\\
Image + Event + Place + PG. & \textbf{19.81} & \textbf{26.49} & 54.70 & 53.92 \\
\midrule 
GT & - & 83.08 & 86.13 & 75.63 \\  
\bottomrule
\end{tabular}
\end{center}
\caption{Generating Inferences using Beam Search vs Nucleus Sampling on the Test set.}
\label{tab:decoding}
\end {table}

\section{Event and Place Generation}
 We report the performance of event and place generation given an image. We try two training schemes with the same model architecture used for generating inferences: 1) train only on event and place, and 2) train on event, place, and inference. The second model is the same model [Image + Event + Place + PG + EP Loss] in Table 2 of the main paper. Note that there are around 10 times more inference sentences than events, meaning the second setup has access to 10 times more data. For fair comparison between the two models, we randomly sample 10\% of the data (Row 2 in Table \ref{tab:event}) and train the second model. 

Table \ref{tab:event} shows the performance of two settings. We report the language metrics, CIDER \cite{cider}, BLEU-4 \cite{bleu}, METEOR \cite{meteor}, and ROUGE \cite{rouge}, vocab size, and sentence length. Overall, we see that the two models perform similarly when the same amount of data are given. CIDER is higher for the first model, while the rest of language metrics are lower. When we use the entire data (All) for the second setup, we see that the improvement is significant for both language metrics and vocab size.

\myparagraph{Inference using Generated Event}
Can the generated event be used as text input to generate the inferences? We use the generated event from Row3 in Table \ref{tab:event} as auxiliary text input and evaluate the quality of inferences. In Table \ref{tab:human_gen} we show human evaluation using the same images and setup in Table 3 of the main paper. Under the section \textit{With Generated Text Input}, we see that the Image + Text model performs better than Text only model, when generated event and place is given as input. However, the scores are lower than the best model without text input (36.0 vs 38.2). Note that this does not indicate that event and place information are not useful. As mentioned in the main paper, the model trained to generate event, place, and inference [Image + Event + Place + PG + EP Loss] performs the best when image is only given as input.

\begin {table}[ht!]
\begin{center}
\setlength\tabcolsep{0pt}
\scalebox{0.85}{
\begin{tabular}{@{}l@{\ \ \ }c@{\ \ }@{}@{}c@{\ \ }c@{\ \ }c@{ \ \ }c@{\ \ }c@{\ \ }c@{\ \ }}
\toprule
\textbf{Training Scheme  } & \textbf{C} & \textbf{B-4} & \textbf{M}   & \textbf{R} &  \textbf{Vocab} & \textbf{Sent Len} \\

\midrule
Image $\rightarrow{}$ Event + Place & 17.61 & 1.85 & 11.78 & 22.62 & 1632 & 9.61 \\
Image $\rightarrow{}$ Event + Place + Inference (10\%) & 15.69 & 2.35 & 12.01 & 23.34 & 1618 & 10.10 \\ 
\midrule
Image $\rightarrow{}$ Event + Place + Inference (All) & 22.97 & 3.47 & 13.21 & 25.23 & 2578 & 9.71 \\ 
\midrule 
GT &  &  &  &  &  3799 & 9.98 \\ 
\bottomrule
\end{tabular}
}
\end{center}
\caption{Event + Place Generation Performance on Test Set. We report the following language metrics: CIDER (C), BLEU-4 (B-4), METEOR (M), and ROUGE (R). We additionally include vocab size and sentence length. See Section E for more details.} 
\label{tab:event}
\end {table}

\begin {table}[t!]
\begin{center}
\scalebox{0.85}{
\begin{tabular}{@{}l@{\ \ } c@{\ \ } c@{\ \ }c@{\ \ }c@{\ \ }c@{\ \ }c@{\ \ }| c@{\ \ }}
\toprule
\textbf{Modalities  } & \textbf{Human} & \textbf{  Human  } & \textbf{  Human  } & \textbf{  Human  } \\

\textbf{} & \textbf{  Before } & \textbf{  Intent  } & \textbf{  After  } & \textbf{  Avg  } \\

\midrule
\it{With Generated Text Input} \\
\midrule
Event + Place & 34.6 & 35.8 & 29.5 & 33.3 \\
Image + Event + Place + PG. & \textbf{38.9} & \textbf{37.5} & \textbf{31.7} & \textbf{36.0} \\
Image + Event + Place + PG + EP Loss. & 37.2 & 32.9 & 30.4 & 33.5 \\

\midrule
\it{With GT Text Input.} \\
\midrule
Event + Place & 54.9 & 52.6 & 42.9 & 50.1  \\
Image + Event + Place + PG  & \textbf{63.36} & \textbf{63.5} & \textbf{56.0} & \textbf{61.0} \\
\midrule
\it{Without Text Input.} \\
\midrule
No Input & 5.3 & 4.9 & 3.5 & 4.6 \\
Image + PG  & 38.2 & 34.8 & 30.3 & 34.4\\ 
Image + Event + Place + PG + EP Loss & \textbf{42.9} & \textbf{36.8} & \textbf{34.8} & \textbf{38.2} \\
\bottomrule
\end{tabular}
}
\end{center}
\caption{\textbf{Generated Inference Results.} Human score for the generated inferences on the Test split. We select 200 random images and generate 5 sentences for each of the three inference type (3000 sentences total). Then, we assign three annotators to determine if each inference sentence is correct, and take the majority vote. Refer to Table 2 and Section 6.2 for model details. We see that the best model using generated event and place as input provides a worse performance than the best model without the text input.} 
\label{tab:human_gen}
\end {table}

\begin{figure}[ht!]%
    \centering
    \includegraphics[height=0.9\textwidth, angle=90]{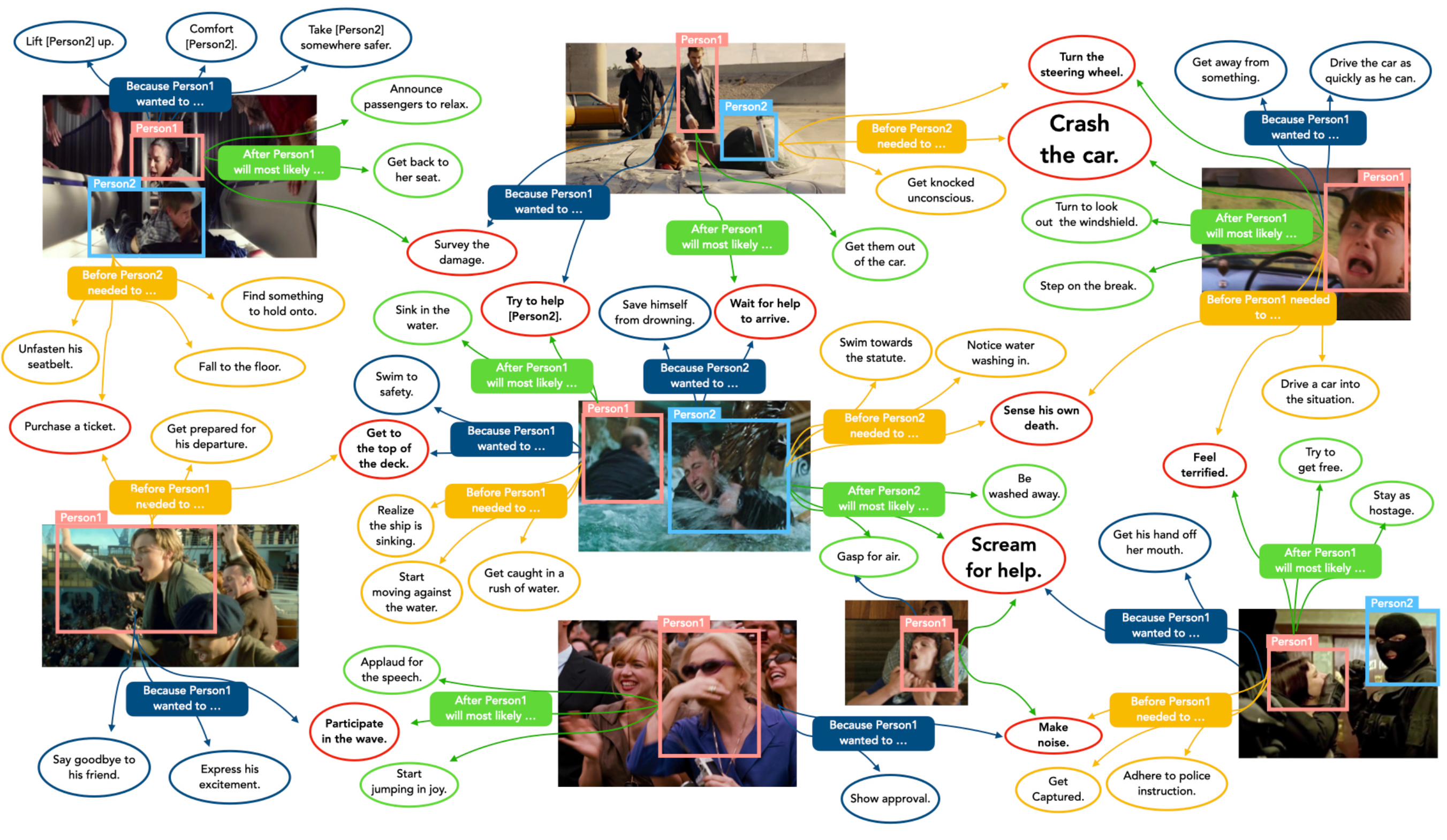} \\ 
   \caption{Overview of our Visual Commonsense Graphs}
    \label{fig:v_comet_viz}
\end{figure}

\clearpage

%
%
\bibliographystyle{splncs04}
\bibliography{biblioLong,egbib}
\end{document}